\documentclass{article}

\usepackage{PRIMEarxiv}

\usepackage{hyperref}       % hyperlinks
\usepackage{url}            % simple URL typesetting
\usepackage{booktabs}       % professional-quality tables
\usepackage{amsfonts}       % blackboard math symbols
\usepackage{nicefrac}       % compact symbols for 1/2, etc.
\usepackage{microtype}      % microtypography
\usepackage{lipsum}
\usepackage{fancyhdr}       % header
\usepackage{graphicx}       % graphics
\graphicspath{{media/}}     % organize your images and other figures under media/ folder

\usepackage{amsfonts,amssymb,amsbsy,latexsym,amsmath,tabulary,graphicx,times,xcolor}

\usepackage[T1]{fontenc}
\usepackage{mathtools}

\usepackage{tikz}
\usepackage[caption=false]{subfig}
\usepackage{float}
\usepackage{hyperref}
\usepackage{xcolor}
\definecolor{darkblue}{rgb}{0, 0, 0.5}
\hypersetup{colorlinks=true,citecolor=darkblue, linkcolor=darkblue, urlcolor=darkblue}

\usepackage{url,multirow,morefloats,floatflt,cancel,tfrupee}
\makeatletter

\AtBeginDocument{\@ifpackageloaded{textcomp}{}{\usepackage{textcomp}}}
\makeatother
\usepackage{colortbl}
\usepackage{xcolor}
\usepackage{pifont}
\usepackage[nointegrals]{wasysym}
\makeatletter

\def\mcWidth#1{\csname TY@F#1\endcsname+\tabcolsep}

\def\cAlignHack{\rightskip\@flushglue\leftskip\@flushglue\parindent\z@\parfillskip\z@skip}
\def\rAlignHack{\rightskip\z@skip\leftskip\@flushglue \parindent\z@\parfillskip\z@skip}

\@ifundefined{etal}{}{}

\usepackage{ifxetex}
\ifxetex\else\if@twocolumn\@ifpackageloaded{stfloats}{}{\usepackage{dblfloatfix}}\fi\fi

\AtBeginDocument{
\expandafter\ifx\csname eqalign\endcsname\relax
\def\eqalign#1{\null\vcenter{\def\\{\cr}\openup\jot\m@th
  \ialign{\strut$\displaystyle{##}$\hfil&$\displaystyle{{}##}$\hfil
      \crcr#1\crcr}}\,}
\fi
}

\AtBeginDocument{%
  \@ifpackageloaded{endfloat}%
   {\renewcommand\efloat@iwrite[1]{\immediate\expandafter\protected@write\csname efloat@post#1\endcsname{}}}{\newif\ifefloat@tables}%
}%

\def\BreakURLText#1{\@tfor\brk@tempa:=#1\do{\brk@tempa\hskip0pt}}
\let\lt=<
\let\gt=>
\def\processVert{\ifmmode|\else\textbar\fi}

\@ifundefined{subparagraph}{
\def\subparagraph{\@startsection{paragraph}{5}{2\parindent}{0ex plus 0.1ex minus 0.1ex}%
{0ex}{\normalfont\small\itshape}}%
}{}

\newcommand\role[1]{\unskip}
\newcommand\aucollab[1]{\unskip}
  
\@ifundefined{tsGraphicsScaleX}{\gdef\tsGraphicsScaleX{1}}{}
\@ifundefined{tsGraphicsScaleY}{\gdef\tsGraphicsScaleY{.9}}{}
\def\checkGraphicsWidth{\ifdim\Gin@nat@width>\linewidth
	\tsGraphicsScaleX\linewidth\else\Gin@nat@width\fi}

\def\checkGraphicsHeight{\ifdim\Gin@nat@height>.9\textheight
	\tsGraphicsScaleY\textheight\else\Gin@nat@height\fi}

\def\fixFloatSize#1{}%\@ifundefined{processdelayedfloats}{\setbox0=\hbox{\includegraphics{#1}}\ifnum\wd0<\columnwidth\relax\renewenvironment{figure*}{\begin{figure}}{\end{figure}}\fi}{}}
\let\ts@includegraphics\includegraphics

\def\inlinegraphic[#1]#2{{\edef\@tempa{#1}\edef\baseline@shift{\ifx\@tempa\@empty0\else#1\fi}\edef\tempZ{\the\numexpr(\numexpr(\baseline@shift*\f@size/100))}\protect\raisebox{\tempZ pt}{\ts@includegraphics{#2}}}}

\AtBeginDocument{\def\includegraphics{\@ifnextchar[{\ts@includegraphics}{\ts@includegraphics[width=\checkGraphicsWidth,height=\checkGraphicsHeight,keepaspectratio]}}}

\DeclareMathAlphabet{\mathpzc}{OT1}{pzc}{m}{it}

\def\URL#1#2{\@ifundefined{href}{#2}{\href{#1}{#2}}}

\def\UrlOrds{\do\*\do\-\do\~\do\'\do\"\do\-}%
\g@addto@macro{\UrlBreaks}{\UrlOrds}

\edef\fntEncoding{\f@encoding}

\makeatother

\newif\ifmultipleabstract\multipleabstractfalse%
\usepackage[normalem]{ulem}
\makeatletter
\newcommand{\xMapsto}[2][]{\ext@arrow 0599{\Mapstofill@}{#1}{#2}}
\def\Mapstofill@{\arrowfill@{\Mapstochar\Relbar}\Relbar\Rightarrow}
\makeatother

\title{The Analysis of Synonymy and Antonymy in Discourse Relations: An interpretable Modeling Approach
}

\author{
  A. Reig-Alamillo\\
  Centro de Investigación en Ciencias Cognitivas \\
  Universidad Autónoma del Estado de Morelos \\
  Cuernavaca, México\\
  \texttt{assela.reig@uaem.mx}\\
  \And
  D. Torres-Moreno \\
  Centro de Investigación en Ciencias \\
  Universidad Autónoma del Estado de Morelos \\
  Cuernavaca, México\\
 \texttt{david.torres@uaem.mx} \\
  \And
  E. Morales-González\\
  Centro de Investigación en Ciencias \\
  Universidad Autónoma del Estado de Morelos \\
  Cuernavaca, México\\
 \texttt{eliseo.moralesgon@uaem.edu.mx}\\
   \And
  M. Toledo-Acosta \\
  Centro de Investigación en Ciencias \\
  Universidad Autónoma del Estado de Morelos \\
  Cuernavaca, México\\
 \texttt{mauricio.toledo@uaem.mx}\\
   \And
  A. Taroni \\
  Centro de Investigación en Ciencias \\
  Universidad Autónoma del Estado de Morelos \\
  Cuernavaca, México\\
  \texttt{antoine.taroni@phelma.grenoble-inp.fr} \\
   \And
  J. Hermosillo-Valadez\thanks{\textit{Corresponding author}} \\
  Centro de Investigación en Ciencias \\
  Universidad Autónoma del Estado de Morelos \\
  Cuernavaca, México\\
  \texttt{jhermosillo@uaem.mx} \\
}

\begin{document}
\maketitle

\begin{abstract}
The idea that discourse relations are construed through explicit content and shared, or implicit, knowledge between producer and interpreter is ubiquitous in discourse research and linguistics. However, the actual contribution of the lexical semantics of arguments is unclear. We propose a computational approach to the analysis of contrast and concession relations in the PDTB corpus. Our work sheds light on the extent to which lexical semantics contributes to signaling explicit and implicit discourse relations and clarifies the contribution of different parts of speech to each of them. This study contributes to bridging the gap between corpus linguistics and computational linguistics by proposing transparent and explainable models of discourse relations based on the synonymy and antonymy of their arguments.
\end{abstract}

% keywords can be removed
\keywords{Concession \and Discourse relation modeling \and Lexical semantic relations}

\section{Introduction}
\label{sec:intro}
The interpretation of a discourse as a content unit, instead of as the mere juxtaposition of independent sentences, is possible by virtue of the existence of coherence or discourse relations. Discourse relations (DRs) are an interdisciplinary object of study: on the one hand, using mainly corpus studies, linguistics has recently contributed to our understanding of how these relations are marked in the discourse%  and processed by the interpreter 
\cite{hoek2021expectations, crible2020, taboada2019space}; %,yung2017psycholinguistic}; % researchers in computational linguistics have focused essentially on parsing the intentional, informational and attentional structures as crucial components of discourse comprehension \cite{el2020analyzing}, \cite{shmueli2019detecting}, \cite{wang-etal-2017-two}; 
on the other hand, research in Natural Language Processing (NLP) and computational linguistics aims chiefly at discourse marker prediction for downstream representation or classification tasks \cite{atwell2021we, bakshi-sharma-2021-transformer, zeldes-etal-2021-disrpt, nie-etal-2019-dissent}. 

%These disciplinary advances share their theoretical object of study and the use of corpus, yet the interdisciplinary dialogue between them is scarce. 
These disciplinary advances share the theoretical object of study and the use of corpus in the endeavor, yet they very rarely translate into an interdisciplinary dialogue between them that would ultimately broaden our understanding of how discourse coherence is constructed and to what extent the linguistically encoded material contributes to the establishing of different types of discourse relations.
 
The contributions of this study are twofold. From a theoretical point of view, it is recognized that linguistically encoded material other than connectives contributes to the signaling of DRs, yet it is unclear how much weight is attributed to lexical semantics in this process and whether different types of DRs behave similarly in this respect. We seek to answer the questions of how much lexical relations (specifically, synonymy and antonymy) contribute to the interpretation of explicit and implicit contrast and concession relations and whether different parts of speech (POS) (a.k.a lexical classes) are of equal importance in signaling these relations, thus contributing to the ongoing debate on the interrelationship between contextual signals and connectives. 

At the same time, from a methodological viewpoint, although very valuable proposals have been put forward to manually encode different types of linguistic signals in corpus \cite{das2019multiple, crible2019domains, crible2022syntax}, much less effort is put in developing corpus analysis methodologies which do not rely, or rely less, on manual analysis and which, instead, take advantage of computational representations of corpus. Hence, we advance towards creating interpretable computational representations, which do not rely on manual coding and which allow us to analyze the lexical-semantic signaling of discourse relations in corpus, thus contributing to provide tools for the benefit of the interface between corpus and computational linguistics.  

The rest of the article is organized as follows. In Section \ref{sec:preguntas} we state the problem and formulate the research questions. In Section \ref{sec:methods} we propose methods for the computational modeling and analysis of semantic synonymy-antonymy signals in discourse relations of contrast.  In Section \ref{sec:results} we present the experimental results after applying our models to the Penn Discourse Treebank-3. We discuss our modeling choices and our results in Section \ref{sec:discusion}, and finally conclude in Section \ref{sec:conclusion}.

\section{Problem Formulation and Research Questions}
\label{sec:preguntas}
In this section, we begin by highlighting the need to understand the contribution of lexical semantics in establishing negative causal relations. We continue with the motivation to propose a new computational approach to their analysis. We conclude this section by posing the research questions and hypotheses. 

\subsection{Discourse Relations and Lexical Semantics}

The processes of establishing suprasentential relations (a.k.a. discourse or coherence relations) between sentences is partially constrained by the linguistic content of the utterance. The most straightforward way in which a specific DR can be linguistically marked or signaled \cite{taboada2019space} is by means of a connective: its absence or presence differentiates between implicit and explicit DRs, respectively. 

Besides connectives, other elements in the explicit linguistic content are assumed to play a role in the interpretation of discourse relations, thus functioning as cues or signals for the inferring process \cite{das2018signalling, crible2019domains, crible2022syntax}. The conceptual meaning encoded in the discourse segments, and specifically the semantic relations between words, are one of such signals guiding the interpretation of a given discourse relation and interacting with connectives \cite{das2019multiple, crible2022syntax}. 
It is, for example, intuitively clear that the semantics of \emph{cold} and \emph{warm} is responsible for the contrast reading in (1). In (2) and (3), the semantic content of \emph{rained} and \emph{dry}/\emph{wet}, together with the interpreter’s world knowledge, would lead to a concessive (2) or a causal (3) discourse relation.

\begin{enumerate}
    \item In New York, it’s cold today; in Mexico City, it is warm.
    \item It rained; the streets are dry.
    \item It rained; the streets are wet.
\end{enumerate}

Negative causal relations (including contrast and concession causal relations) have been defined as a non-basic cognitive relations \cite{sanders1992toward}. In our examples, although the cognitive ability to establish coherence between sentences would tend to assign a concessive interpretation to (2), the absence of an explicit connective (\emph{however}, \emph{but}, \emph{nevertheless}), guiding the interpreter in this inferring endeavor, makes it more cognitively costly than the same endeavor in (3). This difference is consistently found in the psycholinguistic research and corpus studies find that negative causal relations are more often explicitly marked by means of a connective than positive causal relations \cite{spooren2008acquisition, xiang2015reversing, xu2018influence}.

In this study, we are interested in the contribution of lexical semantics and, specifically, synonymy and antonymy, to establishing negative causal relations.
The idea that antonymy can signal contrast is present in the literature. \cite{marcu2002unsupervised} provided evidence of the role of lexical item pairs as clues in building an unsupervised DR classification system, setting the ground to wonder what the role of lexical patterns is in different relations. \cite{spenader2007antonymy} found that antonymy in adjectives seemed to be source of contrast only in the \emph{but} marked pairs of sentences. \cite{feltracco2018lexical} analyzed the role of conceptual opposition ---manually encoded--- in contrast relations, and found a low presence of opposites in the arguments of a contrast relation, and a higher occurrence of opposites when the relation is implicit than explicit (16\% vs 5.2\%). 
These results contribute to picturing the role of antonymy in contrast relations, yet the number of occurrences analyzed is low. More recently, in a corpus study with manually annotated discourse signals, \cite{crible2022syntax} found that semantic relations, mainly antonymy, has a relevant presence in contrast relations (and not in concessive or additive relations), although her study includes only relations with explicit connectives.

The contribution of synonymy to negative causal discourse relations is less straightforward, yet its role in constructing discourse coherence is clearly acknowledged: Synonyms are key ---together with other anaphoric elements--- in establishing topic continuity \cite{Givn1983TopicCI}, hence contributing to local coherence \cite{spooren2008acquisition, taboada2019space}. We posit that synonymy is a useful feature for representing lexical content in discourse as well as a relevant feature that deserves further attention regarding its contribution to different discourse relations \cite{lei2018linguistic}. 

To sum up, the idea that the interpretation of discourse relations is based both on the content explicitly encoded in the discourse and on the common knowledge shared by producer and interpreter, thanks to the human cognitive ability to infer implicit meanings, is pervasive in discourse and linguistic studies. How much weight should be ascribed in this process to the lexical semantics of the arguments is, however, an open question. Likewise, it is unclear to what extent different discourse relations can be characterized and distinguished from other types of discourse relations on the basis of their lexical semantic content, and whether this semantic content also affects the explicit use of a connective. 

In order to contribute to answering these questions while advancing in the dialogue between corpus linguistics and computational methodologies, in this article we propose computational representations of concession and contrast discourse relations that capture the contribution of the semantic conceptual content towards the relation.

\subsection{Constructing Linguistically Informative Computational Representations}

%Regarding the scope of the study that concerns us, 
Linguistic research using corpora to study DRs have greatly contributed to the identification of linguistic signals cooccurring with connectives \cite{crible2022syntax, crible2020, taboada2019space}; nevertheless, these analyses are based on the systematic manual annotation of linguistic signals in the corpora. Our approach wonders what advances can be made, in the same direction, dispensing with manual annotation.
Although many computational approaches have been proposed for the analysis of DRs for predictive and classification purposes \cite{atwell2021we, bakshi-sharma-2021-transformer, roth2014combining, lei2018linguistic, biran2013aggregated, sporleder2008lexical, wellner2006classification}, it is questionable how much these advances have contributed to strengthening the dialogue between NLP and theoretical or corpus linguistics. Indeed, in most cases the features taken into account for the prediction or recognition of a given relation remain unknown to the researcher, and the differences and coincidences between individual types of discourse relations are largely ignored  \cite{lei2018linguistic}.

We believe that the dialogue between linguistics and NLP finds a much more fruitful path in the \emph{identification of linguistic patterns} guiding the interpretation of a certain discourse relation. Other studies have advocated a similar approach on DRs \cite{lei2018linguistic, taboada2013annotation} and in NLP in general  \cite{benamara2017evaluative, boleda2020distributional}. Hence, we propose that lexical semantic relations can be a useful starting point to capture the conceptual meaning of the arguments in contrast and concession DRs, as characterized in the Penn Discourse Tree Bank (PDTB). By using computational models, our work seeks to understand the occurrence of antonymy and synonymy in contrast and concession relations, and to provide knowledge on the patterns of their co-occurrence with discourse connectives.

\subsection{Research Questions and Predictions}

Our work addresses the following research questions, one methodological and three theoretical:
\begin{enumerate}
    \item How can discourse relations be computationally modeled in order to capture the contribution of the lexical semantics to the meaning of the discourse relation? We propose to build and analyze interpretable representations of the lexical content of a DR using synonyms and antonyms from the corpus vocabulary for different POS. This should offer answers to the remaining three questions.
    \item How much do different POS contribute to the representation of contrast and concession DRs? POS differ in the semantic content that they prototypically represent and the syntactic and discourse functions that they play in the text. Determining what POS, or what combination of them, contributes more to the representation of contrast and concession relations will shed light on how lexical semantics and discourse coherence interact. 
    \item Are contrast and concession differentiated using these representations? One of the goals of the linguistic analysis and the computational modeling of DRs is being able to set apart different kinds of relations in a corpus. In this sense, it is of interest to determine whether the proposed representations, based on lexical relations, are useful to differentiate contrast and concession relations. Contrast and concession relations are both included under the tag \textbf{Comparison} in the PDTB corpus. The tag \textbf{Contrast} is used when at least two differences between the two arguments are highlighted, whereas the tag \textbf{Concession} is used when a causal relation expected from one argument is denied. In principle, this distinction would lead to hypothesize that, in contrast relations, the \emph{differences highlighted between Argument 1 and Argument 2} would likely be captured by their conceptual meaning (antonyms and/or synonyms in argument 1 and 2), whereas the difference between expected consequences and actual ones would be less closely tied to the conceptual meaning explicitly expressed in the arguments. Based on this idea, we would expect that the representation of discourse relations based on the lexical content of their arguments would set apart contrast and concession relations.
    \item Do implicit and explicit discourse relations of contrast and concession behave similarly in terms of these representations? Previous studies on discourse relations exclusively analyze either implicit  \cite{sporleder2008lexical} or explicit \cite{crible2022syntax} DRs, or seem to operate on the implicit idea that the features that characterize explicit discourse relations should be the same features characterizing implicit ones \cite{biran2013aggregated}; consequently, in most studies differences are not expected or looked for between the two groups. However, the opposite hypothesis is equally, if not more, plausible taking into account the cognitive process of interpreting coherence: an explicit discourse connective should be expected when the lexical semantic content of the discourse segment contributes less to the discourse relation; in turn, in discourse segments where the lexical semantic content is enough to signal the discourse relation, an implicit connective would be more likely. 
\end{enumerate}

\section{Materials and Methods}
\label{sec:methods}

In order to find some semantic signaling pattern, we propose to build a representation of each argument of a discourse relation using synonyms and antonyms from the corpus vocabulary, grouping them into 4 lexical classes: \emph{nouns}, \emph{adjectives}, \emph{verbs}, and \emph{adverbs}. From these argument models, we propose two ways to analyze semantic signaling patterns by connective type in the corpus: Firstly, by constructing a knowledge graph of DRs using these abstractions; and secondly, by finding synonymy/antonymy relations between these representations. We start by describing the corpus and then the modeling approach.

\subsection{The Corpus: PDTB 3.0}

The Penn Discourse Treebank 3.0 (PDTB3) is a large-scale corpus annotated with information related to discourse structure and discourse semantics \cite[cf.][for details]{webber2019penn}. While there are many aspects of discourse that are crucial to a complete understanding of natural language, the PDTB3 focuses on encoding discourse relations. The PDTB3 adopts the predicate-argument view of discourse relations, where a discourse connective (e.g., \emph{because}) is treated as a predicate that takes two text spans as its arguments. The argument that the discourse connective structurally attaches to is called \textbf{arg}$\mathbf{_2}$, and the other argument is called \textbf{arg}$\mathbf{_1}$. The PDTB3 provides annotations for explicit and implicit discourse relations, where an explicit relation contains an explicit discourse connective.

We will only consider DRs of type contrast and concession, both explicit and implicit. The PDTB3 has a total of 26 contrast discourse connectives, and 47 concession connectives, each being the core of many discourse relations between arguments arg$_1$, and arg$_2$. 

\subsection{Modeling Arguments as Bags-of-Synonyms/Antonyms} 
\label{subsec:representations}

For our purposes, a discourse relation is a \textbf{triplet} (arg$_1$, r, arg$_2$), where arg$_k$ represents either arg$_1$ or arg$_2$ related by the connective $r$. For example:  
\begin{enumerate}
\item ~
\begin{minipage}{0.8\textwidth}
\begin{table}[H]
    \centering
    \begin{tabular}{p{4.8cm} c p{4.8cm}}
          {\small The Manhattan U.S. attorney’s office stressed criminal cases from 1980 to 1987,
        averaging 43 for every 10,000 adults.}   &  {\small but}     & {\small the New Jersey U.S. attorney average 16.}\\ %\hline
        \multicolumn{3}{c}{
        \begin{tabular}{>{\centering}m{4.8cm} c >{\centering}m{4.8cm}}
            arg$_1$ & r & arg$_2$
        \end{tabular}
        }%
    \end{tabular}
\end{table}
\end{minipage}
\end{enumerate}

%The problem now is how to represent these triplets in order to analyze the behavior of these relations in terms of their synonymy/antonymy relationship.

In order to build the representations of arguments, we will consider \textit{two-sided} sets of words. In each of these sets, words in the same side are synonyms, and words in opposite sides are antonyms. We build these sets using Wordnet \cite{wordnet}; in other words, we adopt the WordNet model in the decision of what words stand in a synonymy or an antonymy relation. 
In the next sections, we provide definitions and lay out the procedures we follow to construct representations of discursive arguments. 

\subsubsection{Synonym/Antonym Retrieval Function} Let $\mathcal{D}$ be the corpus of documents; in our setting, each document is a discourse relation from the PDTB3 as in Example (4). Let $\mathcal{V}$ be the vocabulary of $\mathcal{D}$. Given a word $w \in \mathcal{V}$, we consider a function $f$ that performs a query to WordNet and returns a set of synonyms of $w$:  $\text{syn}_\mathcal{V}(w)$, and a set of antonyms of $w$:  $\text{ant}_\mathcal{V}(w)$. These subsets are such that $\text{syn}_\mathcal{V}(w) \ni (w,\text{POS}_w)$, and only words pertaining to $\mathcal{V}$ are actually included in either subset; i.e. $\text{syn}_\mathcal{V}(w)\subset \mathcal{V}$, and $\text{ant}_\mathcal{V}(w)\subset \mathcal{V}$. Notice that any of these subsets may be the empty set. Hence, such a function may be formalized as follows:  

\begin{eqnarray}
    \begin{array}{lccl}
        f:{\kern1em} & \mathcal{V} &\longrightarrow & \big( \mathbb{P}(\mathcal{V}), \mathbb{P}(\mathcal{V}) \big)  \\
        & w & \xmapsto[query]{WordNet}
 &  \big( \text{syn}_\mathcal{V}(w)\subset \mathcal{V}, \text{ant}_\mathcal{V}(w)\subset \mathcal{V} \big)
    \end{array}
\end{eqnarray}
where, $\mathbb{P}(\mathcal{V})$ is the power set of $\mathcal{V}$.

\subsubsection{Bags of Synonyms/Antonyms} Let us consider collections of the form:
\begin{equation}
\label{eq:bopos}
C_i = \Big\{C_i^L \subset \mathcal{V}, C_i^R  \subset \mathcal{V}\Big\}, \text{ for some } i=1,2,\ldots,N.
\end{equation}
We will refer to Expression (\ref{eq:bopos}) as a ``bag-of-synonyms/antonyms'', reminiscent of the bag-of-words model \cite{harris1954distributional}. 
In what follows, we describe the procedure to build these collections. Intuitively, we intend each collection to look like this:
\begin{eqnarray}
C_i = \left\{\begin{array}{c}
     \text{synonyms}  \\
     \text{of} \\
     w
\end{array}  \middle|
\begin{array}{c}
    \text{antonyms}  \\
     \text{of} \\
     w
    \end{array}
\right\}.
\end{eqnarray}
%In this way, we consider a 
In fact, we aim at constructing these sets in such a way that they only contain words with the same grammatical form (i.e. adjectives, nouns, verbs or adverbs). In Figure \ref{fig:examples_sets}, we show some actual examples of these sets, obtained from the PDTB3.

\begin{figure}[H]
\small
    \centering
    \begin{tabular}{cccc}
        $\left\{\begin{array}{c}
            \text{large}  \\
            \text{big} \\
            \cdots
        \end{array}  \middle|
        \begin{array}{c}
            \text{small}  \\
            \text{little} \\
            \cdots
        \end{array}\right\}$ &
        $\left\{\begin{array}{c}
            \text{profit}  \\
            \text{benefit} \\
            \cdots
        \end{array}  \middle|
        \begin{array}{c}
            \text{loss} 
        \end{array}\right\}$ & 
        $\left\{\begin{array}{c}
            \text{complete}  \\
            \text{end} \\
            \cdots
        \end{array}  \middle|
        \begin{array}{c}
            \text{begin} 
        \end{array}\right\}$ & 
        $\left\{\begin{array}{c}
            \text{always}  \\
            \text{ever} \\
            \cdots
        \end{array}  \middle|
        \begin{array}{c}
            \text{never} 
        \end{array}\right\}$ \\
Adjectives & Nouns &  Verbs & Adverbs
    \end{tabular}
    \caption{Actual examples of sets $C_i$ by POS.}
    \label{fig:examples_sets}
\end{figure}

For the sake of clarity, we will make the following abuse of notation $w \in C_i \cup C_j
$ to mean that $w \in C_i^L\cup C_i^R\cup C_j^L\cup C_j^R$.
Let $w_0$ be the first word in $\mathcal{D}$. The construction of these collections is as follows:

\begin{description}
\item[Step 1]  $w_0 \in \mathcal{D}$;
\item[Step 2] $C_1 \leftarrow \Big\{ \text{syn}_\mathcal{V}(w_0), \text{ant}_\mathcal{V}(w_0) \Big\}$;
\item[Step 3] $N \leftarrow 1$;
\item[Step 4] \textsc{for each} $w_k \in \mathcal{D}$ \textsc{repeat} \textbf{Step 5} to \textbf{16} \textsc{until} no more words in $\mathcal{D}$ are found:
\item[Step 5] \textsc{if} $w_k \notin \bigcup_{i=1}^N C_i$ \textsc{then:continue, else: go to} \textbf{Step 9};
\item[Step 6] $C_{N+1}  \leftarrow \Big\{ \text{syn}_\mathcal{V}(w_k), \text{ant}_\mathcal{V}(w_k) \Big\}$;
\item[Step 7] $N \leftarrow N + 1$;
\item[Step 8] \textsc{go to} \textbf{Step 4};
\item[Step 9] either $w_k \in C_j^L$ or $w_k \in C_j^R$, for some $j = 1,2, \ldots, N$;
\item[Step 10] \textsc{if} $w_k \in C_j^L$ \textsc{then: continue, else: go to} \textbf{Step 14};
\item[Step 11] $C_j^L \leftarrow C_j^L \cup \text{syn}_\mathcal{V}(w_k)$;
\item[Step 12] $C_j^R \leftarrow C_j^R \cup \text{ant}_\mathcal{V}(w_k)$;
\item[Step 13] \textsc{go to} \textbf{Step 4};
\item[Step 14] $C_j^L \leftarrow C_j^L \cup \text{ant}_\mathcal{V}(w_k)$;
\item[Step 15] $C_j^R \leftarrow C_j^R \cup \text{syn}_\mathcal{V}(w_k)$;
\item[Step 16] \textsc{go to} \textbf{Step 4};
\end{description}

After parsing the corpus, the resulting sets are manually curated in order to reduce redundancy.  At the end of the process, $m$ ($530$ in our case) sets are obtained, which we order according to their POS. Thus, we define a new set $S_{all}$ in the following way:
\begin{gather*}
S_{all} := \big\{
    \begin{array}{cccc}
        \begin{array}{ccc}
             C_1, & \cdots, & C_{n1},
        \end{array} &
        \begin{array}{ccc}
             C_{n1+1}, & \cdots, & C_{n2},
        \end{array} & 
        \begin{array}{ccc}
             C_{n2+1}, & \cdots, & C_{n3},
        \end{array} & 
        \begin{array}{ccc}
             C_{n3+1}, & \cdots, & C_m
        \end{array}
    \end{array}\big\}\\[-\normalbaselineskip]
  \begin{array}{cccc}
        \hspace{30pt}\underbrace{\kern6em}_{\text{Adjectives}(aj)}
     &
    \hspace{10pt}\underbrace{\kern7em}_{\text{Nouns}(n)}  & 
     \hspace{8pt}\underbrace{\kern7em}_{\text{Verbs}(v)} & 
    \underbrace{\kern7em}_{\text{Adverbs}(av)}
 \end{array}
\end{gather*}

Hence, $S_{all}$ contains the total number of collections, and it should be clear now that we can define disjoint subsets $S_{aj}$, $S_{n}$, $S_{v}$, $S_{av}$ by taking the collections that correspond with the appropriate range of indices in $S_{all}$ for each subset. Thus, now we are able to define 4 subsets as follows:
\begin{eqnarray}
\begin{array}{lcl}
    S_{all-aj}:=S_{all}\setminus S_{aj} \qquad & \qquad & \qquad S_{all-n}:=S_{all}\setminus S_{n} \\
    S_{all-v}:=S_{all}\setminus S_{v} \qquad & \qquad &
    \qquad S_{all-av}:=S_{all}\setminus S_{av} 
\end{array}
\end{eqnarray}

\subsubsection{Modeling Arguments as Bags-of-Synonyms/Antonyms \label{sec:ModR}} We now describe the method to build the representations. In this paper, we only consider contrast and concession DRs, but the method can be applied to all kinds of discourse relations. 

We start by mapping every document in $\mathcal{D}$ to a triplet (arg$^i_1$, r$^i$, arg$^i_2$) ---cf. Example (4)--- for some $i=1,2,\ldots,\rho$, where $\rho$ is the number of discourse relations under study. Let $\mathcal{A}_1$ be the set of all arguments arg$_1$, and $\mathcal{A}_2$ the set of all arguments arg$_2$ under consideration. Consider a function $R$:
\begin{equation}
    R:\kern1em\mathcal{A}_1\cup\mathcal{A}_2 \to \mathbb{Z}^m
\end{equation}
and let $(e_1,e_2,\ldots,e_m)$ be an orthonormal basis for $\mathbb{Z}^m$.
Now, the procedure to construct the representations of discourse relations is:
\begin{description}
\item[Step 1] 
\begin{equation}
\label{eq:alpha}
    \alpha_j(\text{arg}^i_k)=|\text{arg}^i_k \cap C_j^R| - |\text{arg}^i_k \cap C_j^L| \text{  for } j=1,2,\ldots,m;
\end{equation}
\item[Step 2] 
\begin{equation}
\label{eq:R}
    R(\text{arg}^i_k) = \alpha_1(\text{arg}^i_k) e_1+\alpha_2(\text{arg}^i_k) e_2+\cdots+\alpha_m(\text{arg}^i_k) e_m.
\end{equation}
\end{description}
where $|\text{arg}^i_k \cap C_j^{\ast}|$ stands for the number of words of $\text{arg}^i_k$ contained in $C_j^{\ast}$.
Thus, instead of representing arguments by their word content, we represent them based on our bags-of-synonyms/antonyms, sorted by POS. 
%This representation can be read in the following way: the argument $arg^i_k$ does not contain any word belonging to the $C_1$, it contains at least a word in the right hand side of $C_2$, and contains at least 3 words in the left hand side of $C_m$. 

The reader may note that the method above considers all collections within $S_{all}$. We will denote by $R_{\text{all}}$ the representation considering all POS functions. Thus, after having processed each word $w$ in the argument arg$^i_k$, its representation is %the result of the addition and subtraction of $1$'s, which results in 
a vector of positive or negative integers, %denoted by $R(arg_k)$ 
which may look like this:
\begin{equation}
\label{eq:Rex}
    R_{\text{all}}=R(\text{arg}^i_k) = (0, -1, \ldots, 3) \in \mathbb{Z}^m.
\end{equation}

On the basis of the same principle, we can use the subsets $S_{all-aj}, S_{all-n}, S_{all-v}, S_{all-av}$ to obtain other representations made up of only bags-of-synonyms/antonyms belonging to specific POS. In this way, we define the additional representations:

\begin{table}[H]
    \caption{The four additional representations to $R_{\text{all}}$.}
    %\centering
    \renewcommand{\arraystretch}{2}
    \begin{tabular}{c p{7cm}}
    Representation & Meaning \\ \hline
        $R_{\text{all}-\text{aj}}$ & Representation obtained by considering only sets with POS nouns, verbs and adverbs.  \\ \hline
        $R_{\text{all}-\text{n}}$ & Representation obtained by considering only sets with POS adjectives, verbs and adverbs. \\ \hline
        $R_{\text{all}-\text{v}}$ & Representation obtained by considering only sets with POS adjectives, nouns and adverbs. \\ \hline
        $R_{\text{all}-\text{av}}$ & Representation obtained by considering only sets with POS adjectives, nouns and verbs. \\ \hline
    \end{tabular}
    \label{tab:four_representations}
\end{table}

\subsection{A Walk-Through Example}

In this section, we provide a walk-through example of the modeling steps of the arguments. The following is an actual example from the PDTB3 corpus. 
\begin{center}
\begin{description}
\item[arg$_1$:]Japan has climbed up from the ashes of World War II and a gross national product of about 800 per capita to reach the heavyweight class among industrialized nations.
\end{description}
\end{center}
%gross: 1^{AJ}_L, 40^{AJ}_L, 57^{AJ}_L, 199^{AJ}_R
%reach: \{353,VERB,L\}\{357,VERB,L\}\{359,VERB,L\}\{364,VERB,L\}\{369,VERB,L\}\{380,VERB,R\}\{408,VERB,L\}\{416,VERB,R\}\{428,VERB,L\}\{436,VERB,L\}\{464,VERB,L\}\{482,VERB,R\}\{495,VERB,L\}

\noindent In this sentence, the token ``climbed'' appears in the following bags of synonyms/antonyms: in n° $351$ made out of verbs, on the left-hand side; in n° $356$ made out of verbs, on the right-hand side; and in n° $494$ made out of verbs, on the left-hand side. The fact a token may be found in more than one set is due to the fact that the bags of synonyms/antonyms are built on the fly, where the same token may be related to other tokens as a synonym/antonym appearing in a distinct set in an opposite side. However, no token appears twice in the same set, either in the same or in the opposite side. Hence, the information about the precise location of the token is coded as: $\{351^V_L,356^V_R,494^V_L\}$. Thus, in the previous argument, each token for which a synonym/antonym bag was found is labeled with all the bags containing the token. For readability purposes, we provide only a few labels for some of the tokens indicating with ellipses dots that there are other labels in between:  

\begin{description}
\item[arg$_1$:]Japan has $\underset{\{351^V_L,356^V_R,494^V_L\}}{\textrm{climbed}}$ up from the ashes of World War II and a 
$\underset{\{1^{AJ}_L, \cdots, 199^{AJ}_R\}}{\textrm{gross}}$ $\underset{\{2^{AJ}_L, 34^{AJ}_R\}}{\textrm{national}}$ product of $\underset{512^{AV}_L}{\textrm{about}}$ 
800 per capita to $\underset{\{353^{V}_L, 357^V_L, \cdots, 495^V_L\}}{\textrm{reach}}$ the heavyweight $\underset{243^N_L}{\text{class}}$ among industrialized nations.
\end{description}

Now, in order to build the argument's vector representation $R$, we follow steps 1 and 2 of Section \ref{sec:ModR} (equations \ref{eq:alpha} and \ref{eq:R}). Hence, for each labeled token, we add -1 at every coordinate indexed by the bag number if the token was found on the left-hand side, or +1 if the token was found on the right-hand side. Therefore, the previous sentence results in the following (sparse) vector:

$$
\begin{array}{rrrrrrrrrrrrrrr}
     R= [\underbracket{-1}_{\tiny \mathbf{1}}, & \underbracket{-1}_{\tiny \mathbf{2}}, & \cdots, & \underbracket{1}_{\tiny \mathbf{34}}, & \cdots, & \underbracket{-1}_{\tiny \mathbf{40}}, & \cdots, & \underbracket{-1}_{\tiny \mathbf{57}},  & \cdots,& \underbracket{1}_{\tiny \mathbf{199}} , & \cdots,& \underbracket{-1}_{\tiny \mathbf{243}}, & \cdots,& \underbracket{-1}_{\tiny \mathbf{351}},& \cdots,\\
     \underbracket{-1}_{\tiny \mathbf{353}}, &  \cdots, \underbracket{1}_{\tiny \mathbf{356}}, & \underbracket{-1}_{\tiny \mathbf{357}}, &  \cdots,& \underbracket{-1}_{\tiny \mathbf{359}},& \cdots,&
      \underbracket{-1}_{\tiny \mathbf{364}}, & \cdots, & \underbracket{-1}_{\tiny \mathbf{369}}, & \cdots, & \underbracket{1}_{\tiny \mathbf{380}}, & \cdots, & \underbracket{-1}_{\tiny \mathbf{408}},& \cdots, \\
      \underbracket{1}_{\tiny \mathbf{416}},& \cdots,& \underbracket{-1}_{\tiny \mathbf{428}},& \cdots,& \underbracket{-1}_{\tiny \mathbf{436}}, & \cdots,& \underbracket{-1}_{\tiny \mathbf{464}},& \cdots, &
       \underbracket{1}_{\tiny \mathbf{482}}, & \cdots, & \underbracket{-1}_{\tiny \mathbf{494}},& \underbracket{-1}_{\tiny \mathbf{495}},& \cdots,& \underbracket{-1}_{\tiny \mathbf{512}},& \cdots ] 
\end{array}
$$
\noindent where ellipses dots indicate the value $0$ repeated many times.

\subsection{Discourse Relations as Knowledge Graphs}
\label{subsec:discursive-relations}

A knowledge graph is an abstract data structure that represents a network of real-world entities and illustrates the relationship between them. It consists of labelled nodes (also called vertices), which represent the entities, interconnected by links (also called edges), which represent the relationships between the entities. In this way, we can visualize in a relatively simple way the interaction between the entities as a graph structure. In what follows, we will refer to this data structure as a knowledge graph or simply as a graph.

In our setting, we will use graphs to describe the relationship between the arguments of a discourse relation by means of a connective. A first approach is to let each node represent an argument expressed in natural language; i.e. without making any abstraction of its lexical content. For each discourse relation kind and each connective type, we consider that each arg$_1$ is a node joined to its respective arg$_2$ by an edge; we call this configuration a \textbf{stick}. As it is very unlikely that one argument (expressed in natural language) appears exactly in more than one discourse relation, every one of these graphs should look like the one depicted in Figure \ref{fig:grafo-inicial}.  

\def\h{2.5}
\begin{figure}[H]
\centering
\scalebox{0.6}{
\begin{tikzpicture}
% coordinates 
\coordinate (r1a1) at (0,0);
\coordinate (r1a2) at (0,\h);
\coordinate (r2a1) at (2,0);
\coordinate (r2a2) at (2,\h);
\coordinate (r3a1) at (4,0);
\coordinate (r3a2) at (4,\h);
\coordinate (r4a1) at (6,0);
\coordinate (r4a2) at (6,\h);

%% vertices
\draw[fill=black] (r1a1) circle (4pt);
\node[below] at (r1a1) {\Large $\text{arg}^1_1$};
\draw[fill=black] (r1a2) circle (4pt);
\node[above] at (r1a2) {\Large $\text{arg}^1_2$};

\draw[fill=black] (r2a1) circle (4pt);
\node[below] at (r2a1) {\Large $\text{arg}^2_1$};
\draw[fill=black] (r2a2) circle (4pt);
\node[above] at (r2a2) {\Large $\text{arg}^2_2$};

\draw[fill=black] (r3a1) circle (4pt);
\node[below] at (r3a1) {\Large $\text{arg}^3_1$};
\draw[fill=black] (r3a2) circle (4pt);
\node[above] at (r3a2) {\Large $\text{arg}^3_2$};

\draw[fill=black] (r4a1) circle (4pt);
\node[below] at (r4a1) {\Large $\text{arg}^4_1$};
\draw[fill=black] (r4a2) circle (4pt);
\node[above] at (r4a2) {\Large $\text{arg}^4_2$};

% \node at (3.5,\h/2) {${\LARGE \mathbf{\cdots}}$};

%%% edges
\draw[thick] (r1a1) -- (r1a2);
\draw[thick] (r2a1) -- (r2a2);
\draw[thick] (r3a1) -- (r3a2);
\draw[thick] (r4a1) -- (r4a2);
\end{tikzpicture}}
    \caption{Graph depicting 4 discourse relations for some marker as a collection of sticks. The subscript corresponds to either argument 1 or 2 of each relation, and the superscript enumerates the relations. }
   \label{fig:grafo-inicial}
\end{figure}
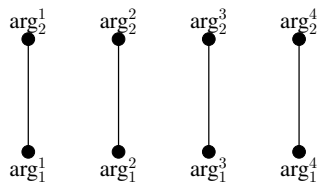

However, this type of graph is not useful for analysing the behaviour of DRs in terms of regularities, or ‘patterns’, of synonymy/antonymy relations, since each entity is related to a single distinct entity. In order to be able to find lexical-semantic patterns between the arguments of DRs, we will use the representations proposed in the previous section. In this case, each node in a network represents the content of each argument in terms of the bags-of-synonyms/antonyms we have defined. 

After obtaining the representation of each argument, we expect to observe a variation in the configuration of the original graph depicted in Figure \ref{fig:grafo-inicial}. This variation could result in the appearance of \emph{central} nodes, which represent arguments that share the same lexical-semantic contents as illustrated in Figure \ref{fig:grafo-ideal}. We say that we may obtain a “richer”, or “more complex” graph, because ramifications from a node may appear, as it is now possible that two, or more, different entities share the same bags-of-synonyms/antonyms: \emph{lexical-semantic similarity patterns appear}.

\def\h{2.5}
\begin{figure}[H]  
\centering 
    \scalebox{0.6}{
    \begin{tikzpicture}
    % coordinates 
    \coordinate (r1a1) at (0,0);
    \coordinate (r1a2) at (-1,1);
    \coordinate (r2a1) at (2,0);
    \coordinate (r2a2) at (2,\h);
    \coordinate (r3a2) at (5,1);
    \coordinate (rna1) at (8,0);
    \coordinate (rna2) at (7,\h);
    %% vertices
    \draw[fill=black] (r1a2) circle (4pt);
    \node[above] at (r1a2) {\Large $R(\text{arg}^1_2)$};
    
    \draw[fill=black] (r2a1) circle (4pt);
    \node[below] at (r2a1) {\Large $R(\text{arg}^1_1)=R(\text{arg}^2_1)=R(\text{arg}^3_1)$};
    \draw[fill=black] (r2a2) circle (4pt);
    \node[above] at (r2a2) {\Large $R(\text{arg}^2_2)$};
    
    \draw[fill=black] (r3a2) circle (4pt);
    \node[above] at (r3a2) {\Large $R(\text{arg}^3_2)$};
    
    \draw[fill=black] (rna1) circle (4pt);
    \node[below] at (rna1) {\Large $R(\text{arg}^4_1)$};
    \draw[fill=black] (rna2) circle (4pt);
    \node[above] at (rna2) {\Large $R(\text{arg}^4_2)$};
    
    % \node at (6,\h/2) {${\LARGE \mathbf{\cdots}}$};
    
    %%% edges
    \draw[thick] (r2a1) -- (r1a2);
    \draw[thick] (r2a1) -- (r2a2);
    \draw[thick] (r2a1) -- (r3a2);
    \draw[thick] (rna1) -- (rna2);
    \end{tikzpicture}}
    \caption{A \textit{rich} knowledge graph for some discourse marker after applying the bag-of-synonyms/antonyms representation to the arguments of 4 relations. Left: the first argument of relations 1, 2, and 3 become identical, reducing to a single node; i.e. they share the same lexical-semantic contents. Right: relationship 4 remains different from the others for the same marker.} 
    \label{fig:grafo-ideal}  
\end{figure}
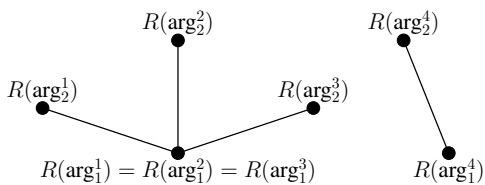  

The reconfiguration of the original graph may lead to two possible extreme cases. The first one is where there are no similarity patterns introduced by our representation. In other words, no two entities share any synonymy/antonymy pattern. The resulting graph is similar to Figure \ref{fig:grafo-inicial}. This situation implies that every entity is different under the representation, that is, they have no common information.
The second case is depicted in Figure \ref{fig:grafo-colapsado}. This situation implies that every entity is the same under the representation. That is, they share all the information in terms of synonymy/antonymy patterns.

%The key step is to consider patterns broad enough so that different entities can have the same representation, yet not so broad that all entities have the same representation. 

\def\h{2.5}
\begin{figure}[H]  
\centering 
\subfloat[A single stick.]{
    \scalebox{0.6}{
    \begin{tikzpicture}
    % coordinates 
    \coordinate (r1a1) at (0,0);
    \coordinate (r1a2) at (0,\h);
    %% vertices
    \draw[fill=black] (r1a1) circle (4pt);
    \node[below] at (r1a1) {\Large $R(\text{arg}^1_1)=...=R(\text{arg}^n_1)$};
    \draw[fill=black] (r1a2) circle (4pt);
    \node[above] at (r1a2) {\Large $R(\text{arg}^1_2)=...=R(\text{arg}^n_2)$};
    %%% edges
    \draw[thick] (r1a1) -- (r1a2);
    \end{tikzpicture}
    }
}\qquad
\subfloat[A single point.]{
    \scalebox{0.6}{
    \begin{tikzpicture}
    % coordinates 
    \coordinate (all) at (1,\h/2);
    %% vertices
    \draw[fill=black] (all) circle (4pt);
    \node[above] at (all) {\Large $R(\text{arg}^1_2)=...=R(\text{arg}^n_2)$};
    \node[below] at (all) {\Large $R(\text{arg}^1_1)=...=R(\text{arg}^n_1)$};
    %%% edges
    % \draw[thick] (r1a1) -- (r1a2);
    \end{tikzpicture}
    }
}
\caption{Two representations for which many entities share the same bag-of synonyms/antonyms.} 
\label{fig:grafo-colapsado}  
\end{figure}

\subsection{Measuring Inter-Relation Synonymy/Antonymy Patterns Between Arguments}
\label{subsec:graph-metrics}

In terms of graphs, one difference between the three situations described above is the presence of nodes with ramifications. These branching nodes can be characterized in terms of high values of \textbf{centrality} measures \cite{bonacich2007some}. 
In graph analysis, centrality is a very important concept for identifying relevant nodes in a graph; it addresses the question: ``What characterizes an important vertex?''. One important centrality measure is the \textbf{eigenvector centrality} (also called eigencentrality) \cite{GOLBECK201325}, which is a measure of the influence of a node in a graph. The presence of important nodes hints the complexity, or richness, of a graph. For example, in the graph of Figure \ref{fig:grafo-ideal}, there is a node with a high centrality value at the left-hand side. 

In our setting, centrality measures allow us to know if there is any node with greater relevance than others, in terms of concentrating a lexical-semantic pattern, thus becoming an argument that links several others. In this way, measuring the importance of a node is equivalent to assessing the relevance of an argument in terms of capturing a lexical-semantic pattern repeated across the corpus in a given type of discourse relation (contrast or concession, in this article). 

In order to quantify the variations in the configuration of graphs, \emph{attributable to lexical-semantic similarity patterns}, we define a centrality-based metric, denoted by $\varphi$, which is \emph{the ratio between the maximum node eigenvector centrality and the average node eigenvector centrality of a graph}. Thus, given a graph $G(V,E)$, let $x_v$ be the node eigenvector centrality of node $v$. Hence,
\begin{equation}
\label{eq:phi}
    \varphi = \frac{\text{max}\{x_v | \forall v \in V\}}{\text{mean}\{x_v | \forall v \in V\}}
\end{equation}

\begin{figure}[hptb!]
    \centering
    \includegraphics[width=0.8\linewidth]{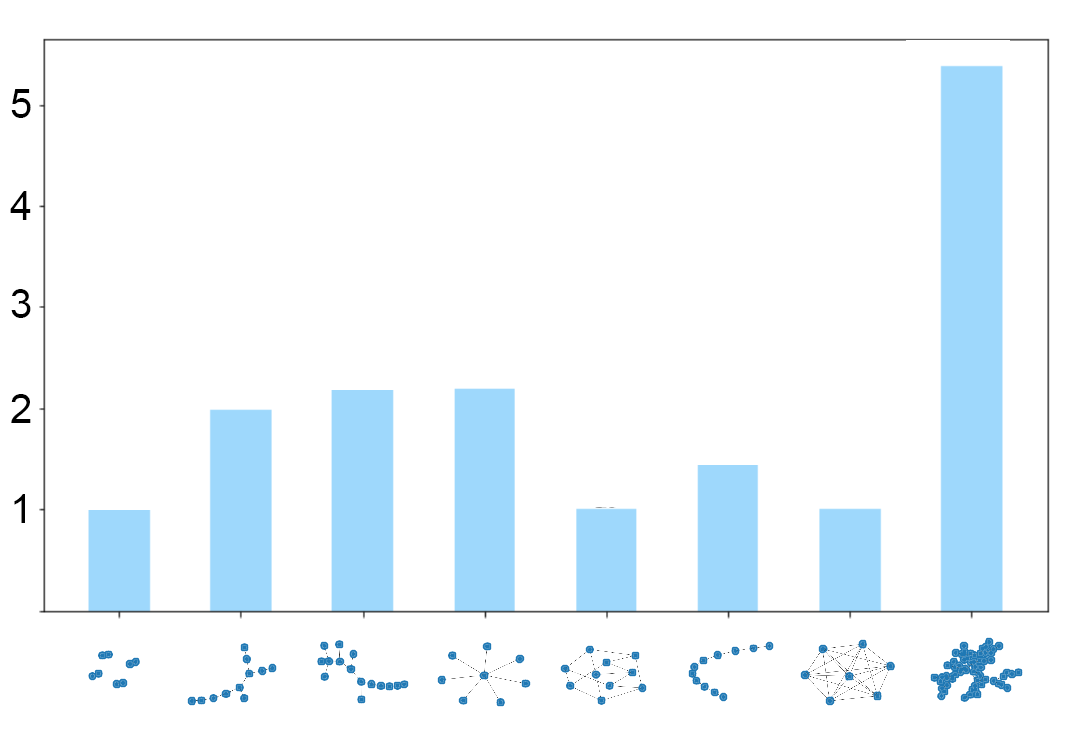}
    \caption{The value of the metric $\varphi$ for several graphs $G$.}
    \label{fig:metric_for_graphs}
\end{figure}

In Figure \ref{fig:metric_for_graphs}, we show several examples of graphs along with their respective value of $\varphi$. The more nodes with several ramifications in a graph, the higher the value of the metric $\varphi$. Another way to read this figure is by observing that the higher the complexity (irregularity), the higher the value of $\varphi$.

In our work, we analyze the graphs of explicit and implicit discourse connectives, using the representation $R_\text{all}$, and observe how these graphs change when we replace $R_\text{all}$ by the representations $R_{\text{all}-\text{aj}}$, $R_{\text{all}-\text{av}}$, $R_{\text{all}-\text{v}}$ and $R_{\text{all}-\text{n}}$. The intention is to analyze the effect of a missing POS on the reconfiguration of the graph of $R_\text{all}$. We will use the metric $\varphi$ to perform the analysis.
%In this paper, we analyze the graphs of explicit and implicit discourse connectives, using the representation $R_\text{all}$, and observe how these graphs change when we replace $R_\text{all}$ by the representations $R_{\text{all}-\text{aj}}$, $R_{\text{all}-\text{av}}$, $R_{\text{all}-\text{v}}$ and $R_{\text{all}-\text{n}}$. We will use the metric $\phi$ to do this analysis.

\subsection{Finding Intra-Relation Synonymy/Antonymy Patterns}
\label{subsec:rep-syn-ant}

In these subsection, we describe how we use the representations defined in Subsection \ref{subsec:representations} to discover the presence of synonymy and antonymy between arguments in a discourse relation. Consider the following example extracted from the manual of the corpus PDTB:
\begin{enumerate}
\item ~
\begin{minipage}{0.8\textwidth}
\begin{table}[H]
    \centering
    \begin{tabular}{p{4.8cm} c p{4.8cm}}
          {\small After all, gold prices usually \textbf{soar} when \textbf{inflation} is high. Utility stocks,}   &  {\small on the other hand,}     & {\small \textbf{thrive} on \textbf{disinflation} $\cdots$}\\ %\hline
        \multicolumn{3}{c}{
        \begin{tabular}{>{\centering}m{4.8cm} c >{\centering}m{4.8cm}}
            ($\text{arg}_1$) & (marker) & ($\text{arg}_2$)
        \end{tabular}
        }%
    \end{tabular}
\end{table}
\end{minipage}
\end{enumerate}

In this case, we can observe the presence of synonymy and antonymy between $\text{arg}_1$ and $\text{arg}_2$ (\emph{soar} is synonymous with \emph{thrive}, and \emph{inflation} is antonymous with \emph{disinflation}). We can detect these \emph{intra-relation synonymy/antonymy matching patterns} using the representations previously defined. 

Consider a discourse relation ($\text{arg}_1$, $r$, $\text{arg}_2$) for some explicit or implicit connective $r$.  Now, consider the representation of each argument, $R(\text{arg}_1)$ and $R(\text{arg}_2)$, given by

$$R(\text{arg}_1) = (a_1,\dots, a_K),\;\;\;\; R(\text{arg}_2) = (b_1,\dots, b_K)$$

These representations can be any of the 5 representations previously defined. The element-wise product of two representations is:

    $$R(\text{arg}_1) \ast R(\text{arg}_2) = (a_1 b_1 , \dots, a_K b_K ).$$

Observe that a positive value ($a_i b_i > 0$) means that, both $a_i$ and $b_i$ have the same sign, that is, \emph{both $arg_1$ and $arg_2$ have synonyms from the same group}. On the other hand, $a_ib_i < 0$ means that, \emph{either $arg_1$ or $arg_2$ have antonyms from the same group}. Hence, after computing $R(\text{arg}_1) \ast R(\text{arg}_2)$, we count the number of positive and negative components, and denote them by $n_{syn}$ and $n_{ant}$ respectively. These counts stand for \emph{the number of synonymy and antonymy matches between $arg_1$ and $arg_2$}, respectively. For example, if $R(\text{arg}_1)=(-2,0,1,0)$, and $R(\text{arg}_2)=(1,-1,1,0)$, then $R(\text{arg}_1) \ast R(\text{arg}_2)=(-2,0,1,0)$ and therefore, $n_{\text{syn}}=1$ and $n_{\text{ant}}=1$.

The pair $\left(n_{\text{syn}},n_{\text{ant}}\right)$ gives a 2-dimensional representation of the discourse relation $(\text{arg}_1,r,\text{arg}_2)$ in terms of the synonymy/antonymy between the arguments. %The first component quantifies the synonymy between arguments, while the second component quantifies the antonymy between arguments.
In the example (5) 
%of Table \ref{tab:example-syn-ant}, 
we would have $n_\text{syn}=1$ and $n_\text{ant}=1$. Therefore, these counts yield a representation for this discourse relation, which is the point $(1,1)$. %This particular relation will contribute to the explicit sub-cell of the cell corresponding to the point $(1,1)$.

%In this section we present the results obtained after applying the methods describe in Section \ref{sec:methods}. 

\section{Corpus Analysis Using the Proposed Representations}
\label{sec:results}
In this section we present the analysis of contrast and concession DRs in the PDTB3 corpus using our representations. We first introduce an arbitrary classification of connectives, based on their frequency of occurrence. We then show some examples of the graphs obtained for some representative connectives, followed by the analysis relative to the $\varphi$ metric for all representations for all connectives in the corpus, in the form of bar charts. This analysis provides information on the inter-relation lexical patterns found in the whole set of contrast and concession relations in the corpus. We conclude this section by visualizing a comparative analysis of the counts of intra-relation synonymy/antonymy patterns between arguments for contrast and concession DRs.

\subsection{Classification of Contrast and Concession Connectives in the PDTB3}

As mentioned, the PDTB3 has a total of 26 contrast discourse connectives, and 47 concession connectives.  Recall that each of these connectives (r) can be the core of many discourse relations ($\text{arg}_1$,r, $\text{arg}_2$). %As described in the previous section, each of these discourse relations is modeled by a triplet $(\text{arg}_1,r,\text{arg}_2)$, where $r$ refers to the connective.
After the $R_{\text{all}}$ representation, some triplets are discarded because they do not include any of the sets. Thus, out of the total number of connectives, we retain 24 contrast connectives and 42 concession connectives. In other words, our bag-of-synonymy-antonymy representation allowed us to keep $92\%$ of the contrast relations and $89\%$ of the concession relations in PDTB3.

We wish to quantify the complexity of the graphs in function of the frequency of the discourse relations per connective. For example, the connective ``but'' appears in many relations, and ``still'' appears in way fewer. We want to avoid a bias in the analysis due to the frequency of the connectives, and rather try to identify whether the linguistic features of synonymy/antonymy are responsible for the properties of these graphs. To accomplish this, we separate in two classes the connectives according to the number of discourse relations for each connective. To accomplish this, we separate in two classes the connectives according to the number of discourse relations for each connective. These classes are described in Table \ref{tab:groupsABC}. 

\begin{table}[H]
    \caption{We divide the connectives in two classes according to the number of discourse relations in each connective.}
    \underline{$\hspace{\textwidth}$}
    \renewcommand{\arraystretch}{1.5}
    \begin{tabular}{ccl}
    Class & Definition & \hspace{5em}Some examples \\ \hline
        A & Connectives with, at least, 100 instances. & but, however, although, by comparison,\\ & & though, while, yet, by contrast\\ \hline % triangulo
        B & Connectives with less than 100 instances & even though, nevertheless, whereas, \\ & & in contrast,  even if, on the other hand, \\ & & without, and, on the contrary, despite, \\ & & in fact, even so, if\\ \hline % rectángulos
        % Group C & Connectives with less than 14 instances. \\ \hline  % circulo
    \end{tabular}
    \label{tab:groupsABC}
\end{table}

\subsection{Representative Knowledge Graphs}

\begin{figure}[H]  
\centering 
\subfloat[$R_{\text{all}}$]{
\scalebox{0.55}{
    \includegraphics[width=0.8\linewidth]{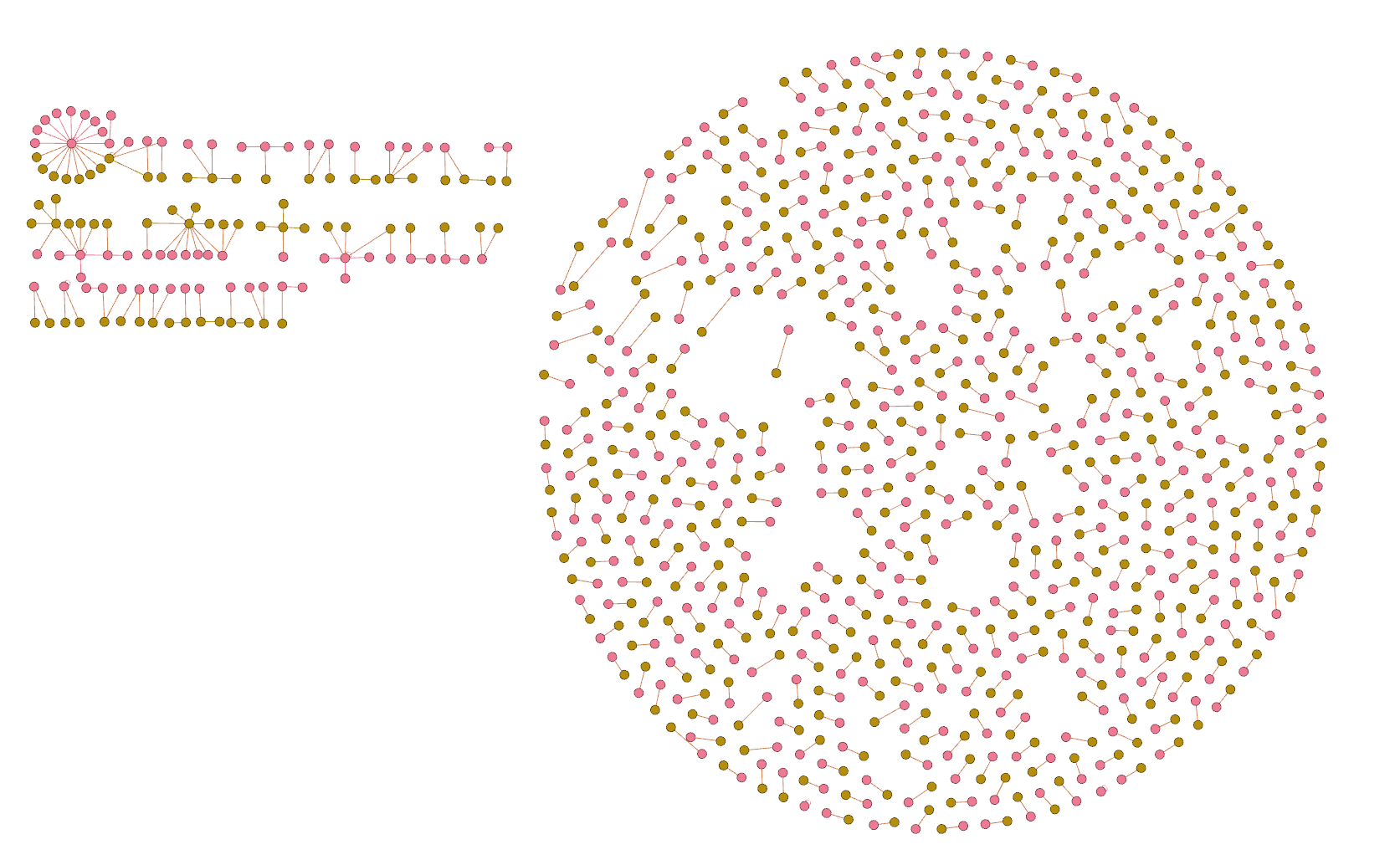}
}   \label{fig:reps-but-contraste-all}
}
\subfloat[$R_{\text{all}-\text{aj}}$]{
\scalebox{0.55}{
    \includegraphics[width=0.8\linewidth]{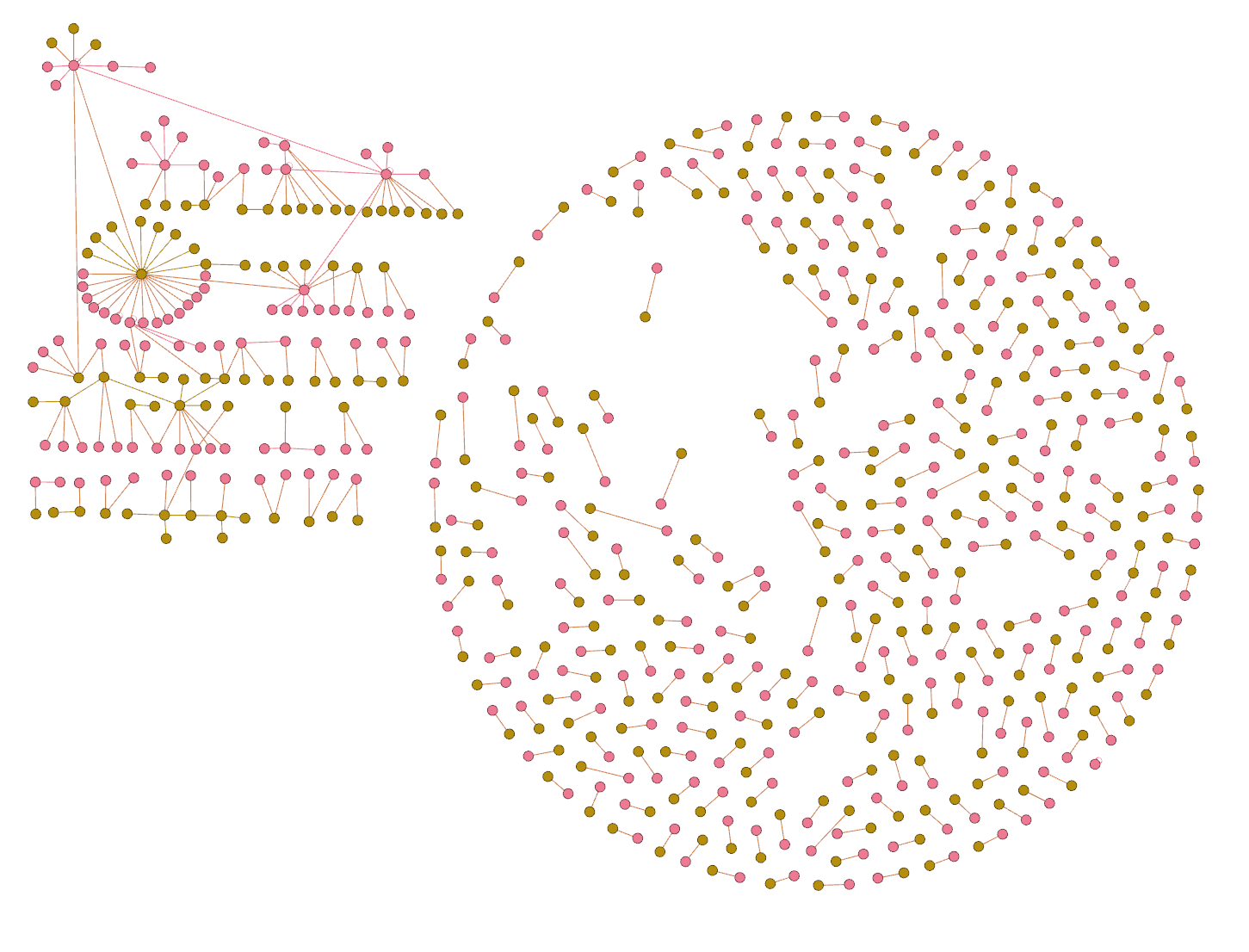}
}   \label{fig:reps-but-contraste-noadj}
}
\caption{Representations of all contrast discourse relations with the connective \textit{but}. We use the representations $R_\text{all}$ and $R_{\text{all}-\text{aj}}$ to show the influence of adjectives.} 
\label{fig:reps-but-contraste}
\end{figure}

\begin{figure}[H]  
\centering 
\subfloat[$R_{\text{all}}$]{
\scalebox{0.55}{
    \includegraphics[width=0.8\linewidth]{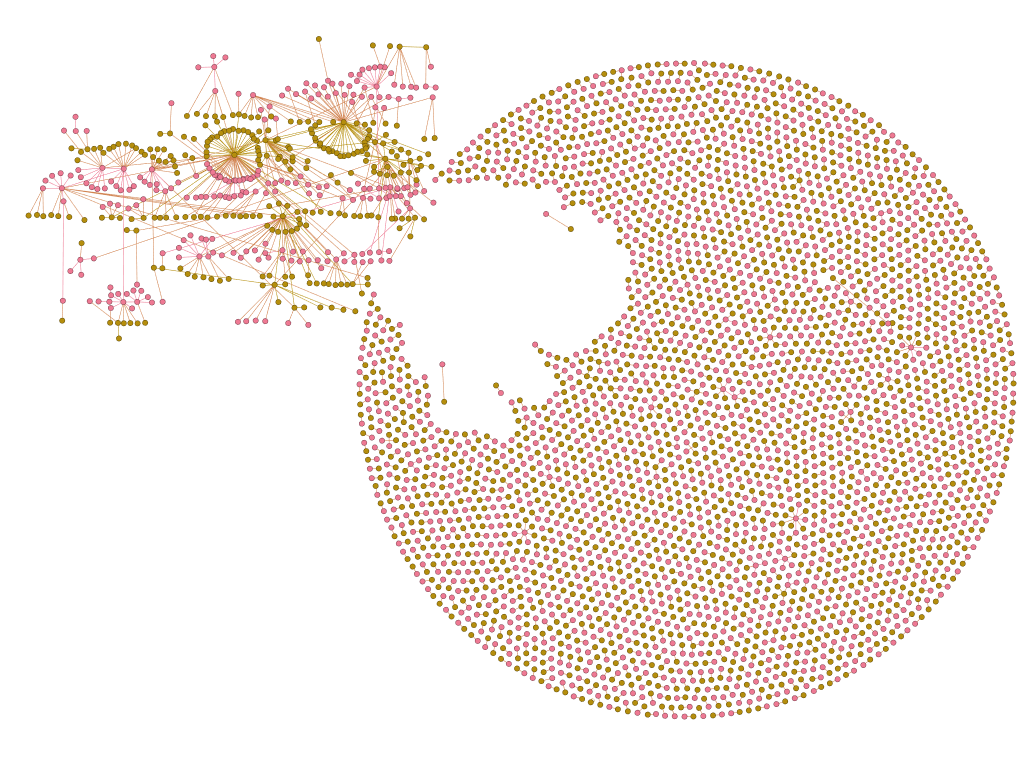}
}    \label{fig:reps-but-concesion-all}
}
\subfloat[$R_{\text{all}-\text{aj}}$]{
\scalebox{0.55}{
    \includegraphics[width=0.8\linewidth]{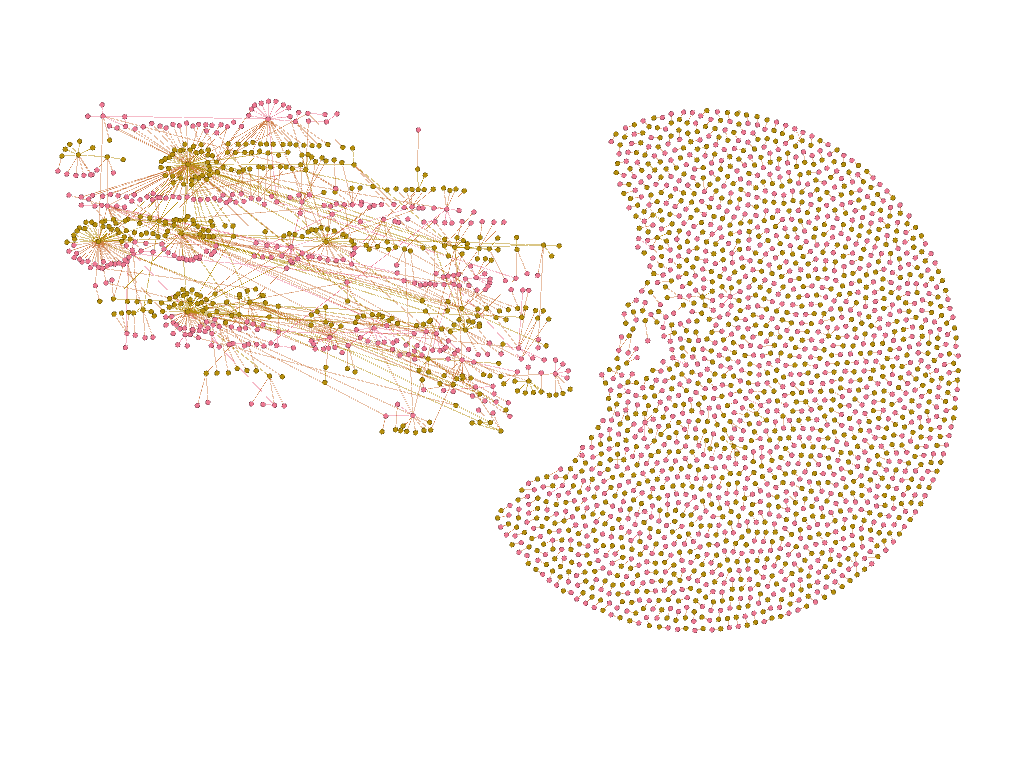}
}    \label{fig:reps-but-concesion-noadj}
}
\caption{Representations of all concession discourse relations with the connective \textit{but}.We use the representations $R_\text{all}$ and $R_{\text{all}-\text{aj}}$ to show the influence of adjectives.} 
\label{fig:reps-but-concesion}
\end{figure}

To illustrate the effect of our representations, we show in Figures \ref{fig:reps-but-contraste} and \ref{fig:reps-but-concesion} the graphs of $R_\text{all}$ (left-hand side) and $R_{\text{all}-\text{aj}}$ (right-hand side) for the connective ``but''.
%s of Table \ref{tab:prototypical-connectives}. %This will allow us to give examples of the behaviour of the graphs when we exclude certain POS subsets, in this case, adjectives. 
%The changes in the configuration of the graphs may shed some light on the importance/weight carried by the missing POS subset. These examples might also give an account of how the metric $\phi$, defined in Section \ref{sec:methods}, reflects the changes in the configuration of the graphs.
%
%These figures show, on the left side, the $R_\text{all}$ graph and on the right side the $R_{\text{all}-\text{aj}}$ graph. 
We have sectioned each graph in order to observe two types of interconnections between nodes. In each graph, one can observe, on the one hand, a collection of sticks, that is, a pair of nodes connected by an edge, distributed in a disk; on the other hand, on one side of the disk, one can observe nodes with branches. % These ramifications can present different configurations, giving rise to different degrees of complexity, as illustrated in Figure \ref{fig:metric_for_graphs} of section \ref{sec:methods}.
%
%In these graphs, we observe the presence of nodes with more branches when 
In both figures we use the representation $R_\text{all}$ and $R_{\text{all}-\text{aj}}$ to show the influence of adjectives.%; i.e. without adjectives. In general, as we move from representation $R_\text{all}$ to representation $R_{\text{all}-\text{aj}}$, we observe the presence of more nodes with branches, except perhaps the connective "still". This means that, by not considering adjectives in the representation, more arguments share the same synonymy/antonymy patterns. In other words, adjectives seem to provide enough discriminating information to make arguments distinct from each other. These changes in the configuration of the nodes can be quantified by our metric $\phi$, as will be shown in the next subsection.
%%%%%%%%%%%%%%%%%%%%%%%%%%%%% CAMBIAR %%%%%%%%%%%%%%%%%%%%%%%%%%

The branching phenomenon gives rise to different degrees of complexity, as explained in Section \ref{sec:methods}. In the graphs above, as we move from representation $R_\text{all}$ to representation $R_{\text{all}-\text{aj}}$ (i.e., without adjectives), we observe the presence of more nodes with branches. 
%, except perhaps the connective ``still''. 
This means that, by not considering adjectives in the representation, more arguments share the same synonymy/antonymy patterns. These changes in the configuration of the nodes can be quantified by our metric $\varphi$, as will be shown in the next subsection.

\subsection{Measuring Inter-Relation Synonymy/Antonymy Patterns Between Arguments: Assessing Graph Complexity}

\begin{figure}[hptb!]
    \centering
    \includegraphics[width=0.9\linewidth]{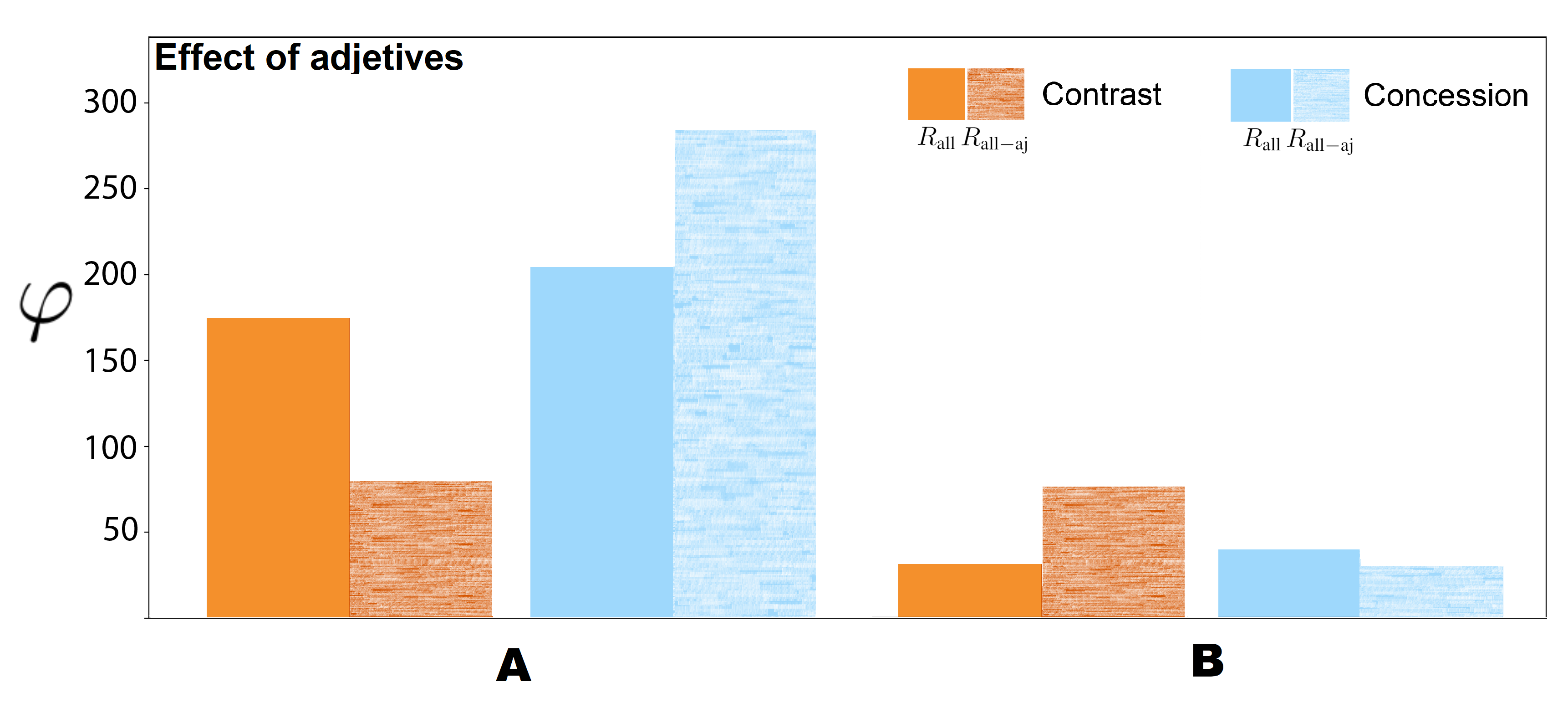}
    \caption{The representation $R_\text{all}$ against the representation $R_{\text{all}-\text{aj}}$.}
    \label{fig:phi_all_vs_no-aj}
\end{figure}

\begin{figure}[hptb!]
    \centering
    \includegraphics[width=0.9\linewidth]{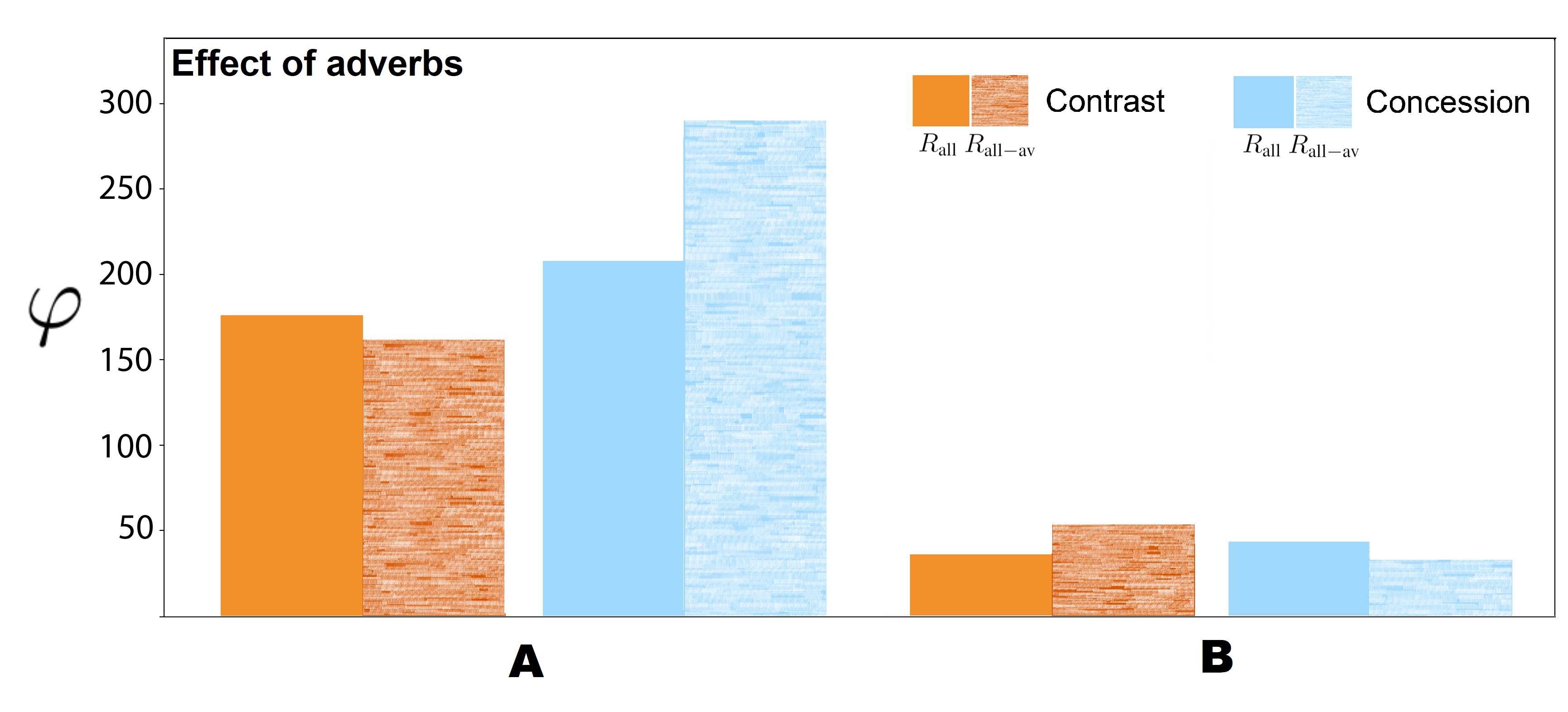}
    \caption{The representation $R_\text{all}$ against the representation $R_{\text{all}-\text{av}}$.}
    \label{fig:phi_all_vs_no-av}
\end{figure}

\begin{figure}[hptb!]
    \centering
    \includegraphics[width=0.9\linewidth]{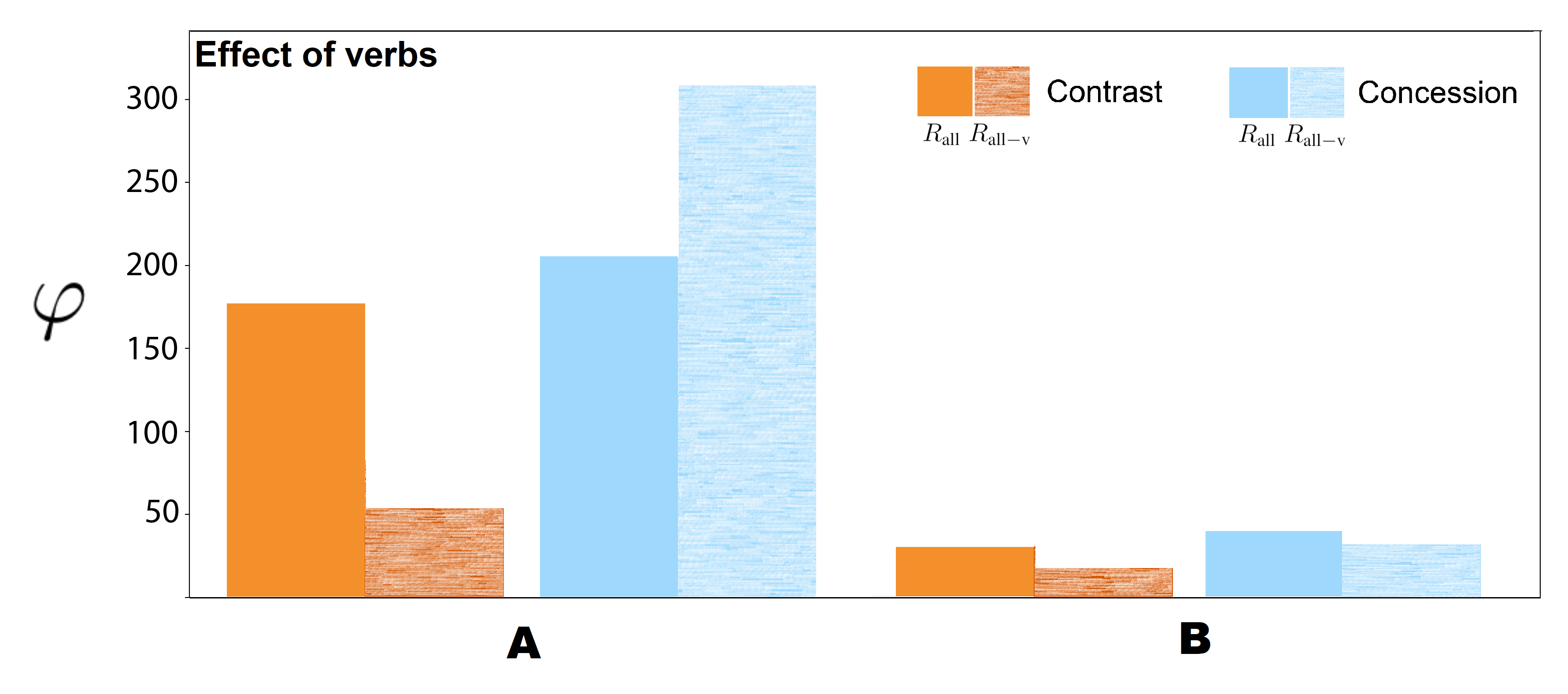}
    \caption{The representation $R_\text{all}$ against the representation $R_{\text{all}-\text{v}}$.}
    \label{fig:phi_all_vs_no-v}
\end{figure}

\begin{figure}[hptb!]
    \centering
    \includegraphics[width=0.9\linewidth]{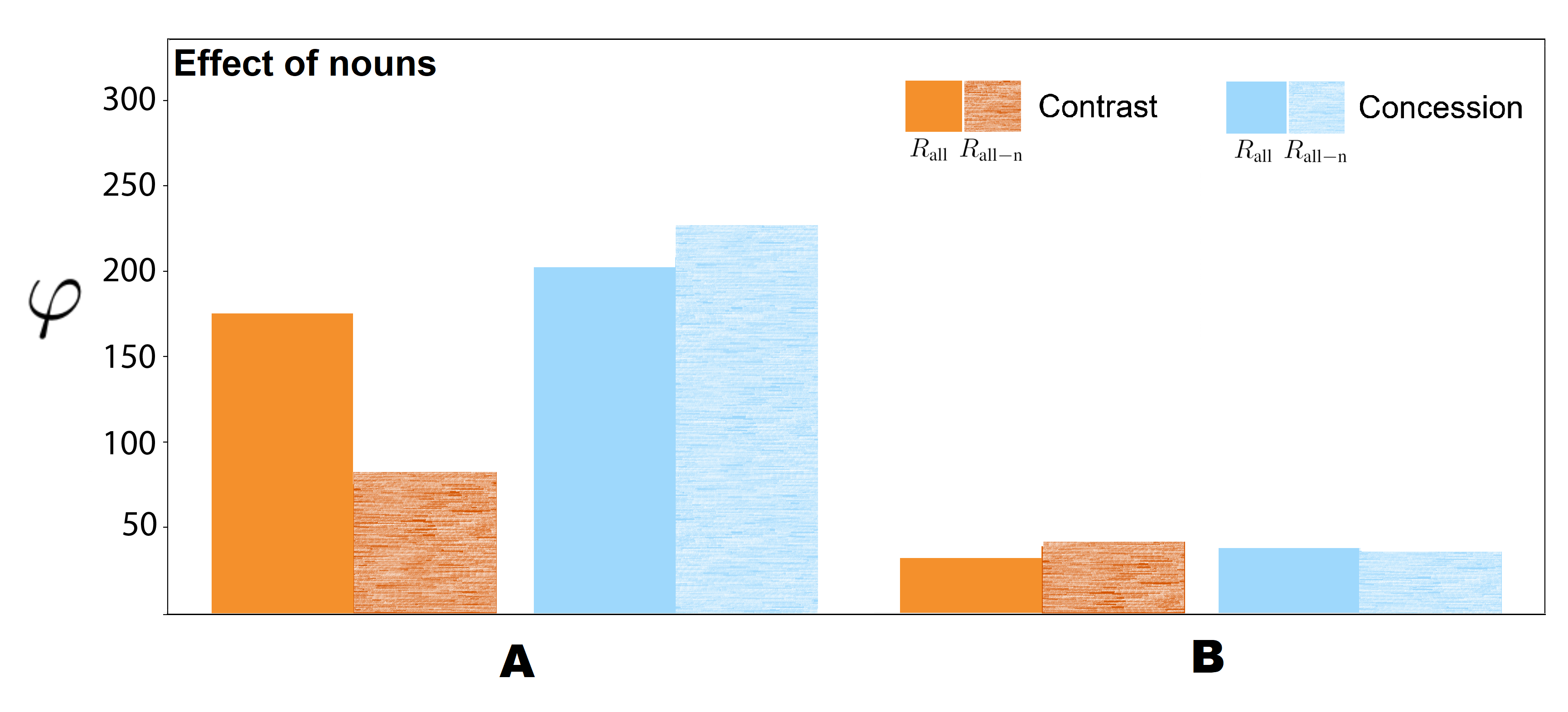}
    \caption{The representation $R_\text{all}$ against the representation $R_{\text{all}-\text{n}}$.}
    \label{fig:phi_all_vs_no-n}
\end{figure}

Now, we turn our attention to the general situation, %but instead of showing all knowledge graphs for all connectives, and all five representations, we look 
by looking at the values of the metric $\varphi$ %instead. We will look at the changes in the values of $\phi$ 
as we move from representation $R_\text{all}$ to representations $R_{\text{all}-\text{aj}}$, $R_{\text{all}-\text{av}}$, $R_{\text{all}-\text{v}}$ and $R_{\text{all}-\text{n}}$. We do this in each of the two classes A (connectives with high frequency of occurrence) and B (connectives with low frequency of occurrence). % 
Recall that the higher the value of $\varphi$, the graph contains nodes with more branches. 
In Figures \ref{fig:phi_all_vs_no-aj}-\ref{fig:phi_all_vs_no-n}, we show the mean of the value of $\varphi$ in each of the two classes A, B. % for representation $R_\text{all}$ and each of the four representations. 
 We show both the contrast (orange bars) and concession (blue bars) relations.

\subsection{Quantifying Synonymy/Antonymy Relationships Between Arguments within a Discourse Relation}

\begin{figure}[H]  
\centering
\scalebox{0.53}{
    \hspace*{-2em}
    \includegraphics[scale=0.40]{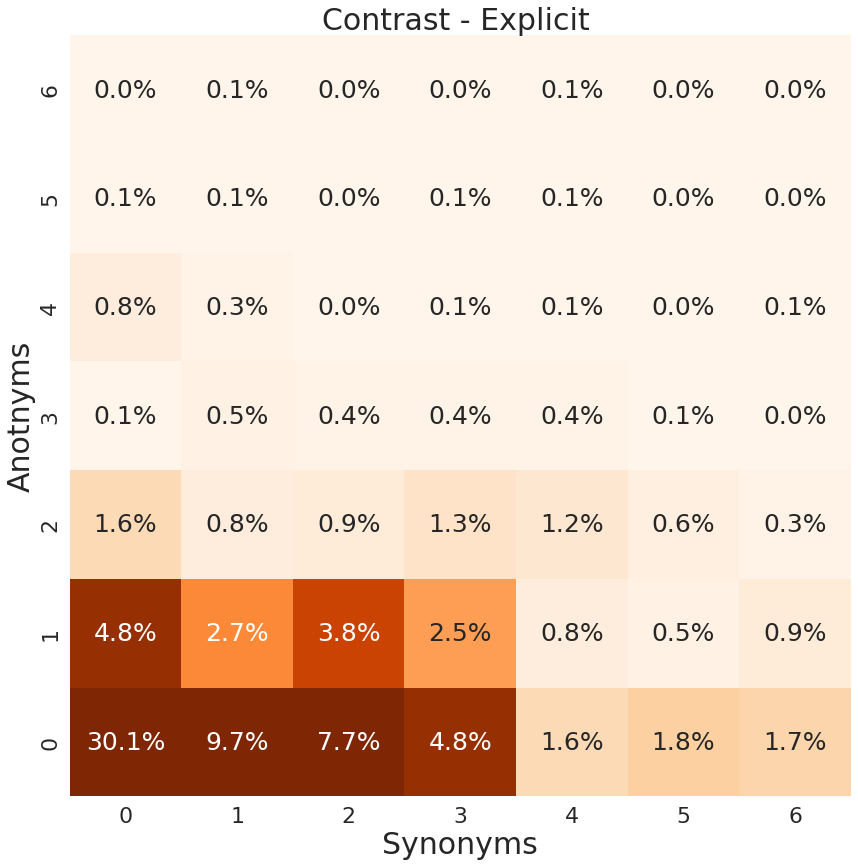}
    \includegraphics[scale=0.40]{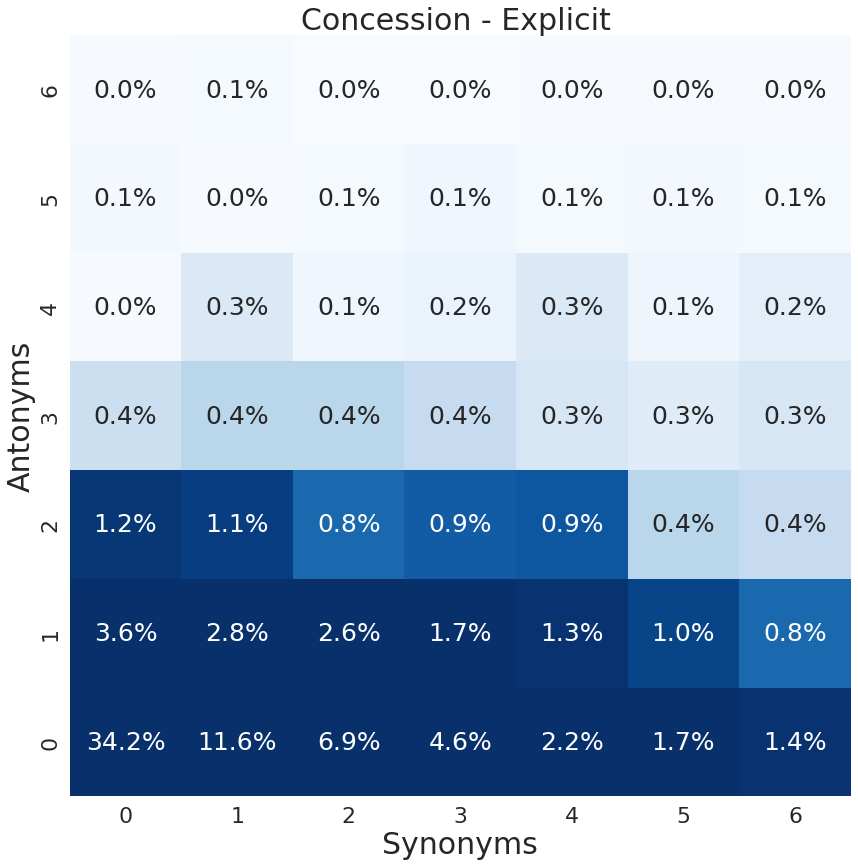}
}    \label{fig:Relation porcentage1}
\caption{Proportion of antonymy and synonymy between arguments in explicit DRs of contrast and concession.} 
\label{fig:producto-concesion-contrasteY}
\end{figure}

\begin{figure}[H]  
\centering
\scalebox{0.53}{
    \hspace*{-2em}
    \includegraphics[scale=0.40]{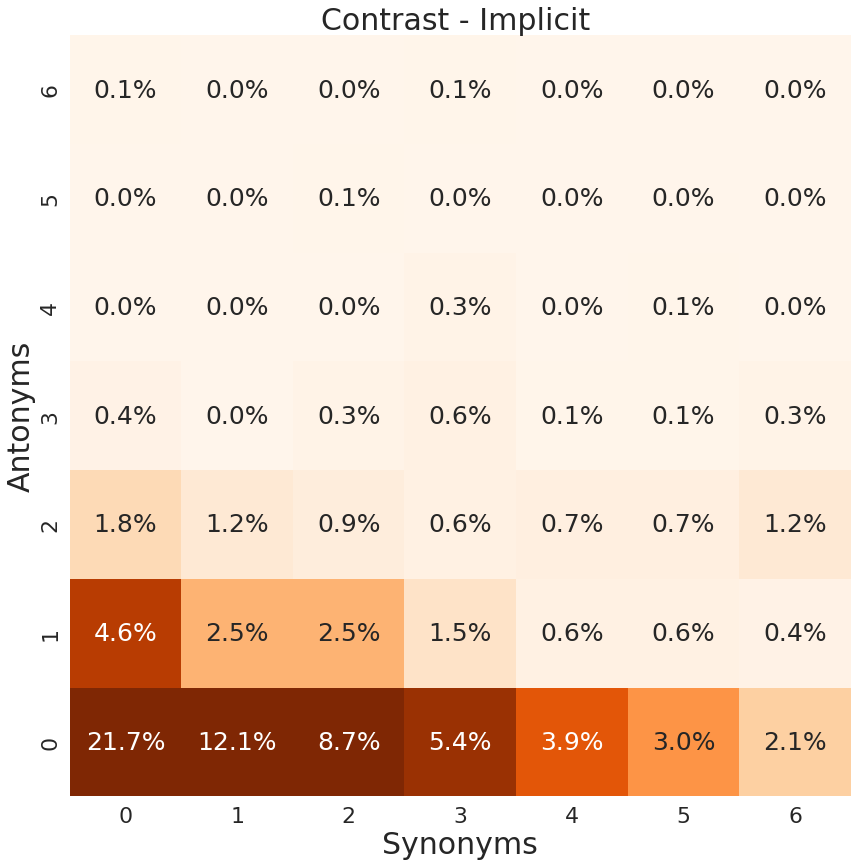}
    \includegraphics[scale=0.40]{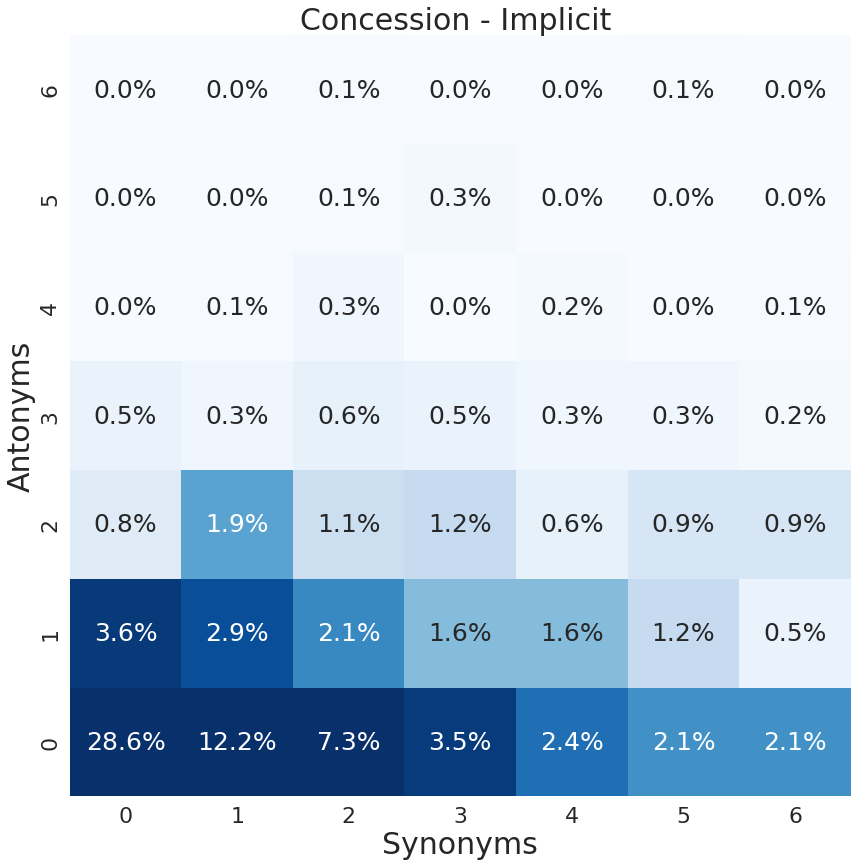}
}    \label{fig:Relation porcentage2}
\caption{Proportion of  antonymy and synonymy between arguments in implicit DRs of contrast and concession. } 
\label{fig:producto-concesion-contrasteX}
\end{figure}

%Regarding the quantification of synonymy and antonymy between arguments in a discourse relation, we use the 2-dimensional representations and visualization described 
We now quantify the number of intra-relation synonymy-antonymy matching correspondences between arguments ($\text{arg}_1$ and $\text{arg}_2$ in a given DR), using the 2-dimensional representations described in Subsection \ref{subsec:rep-syn-ant}. In order to visualize these points, we use heat maps. Each cell in the heat map represents a coordinate $(n_{syn},n_{ant})$. Figure \ref{fig:producto-concesion-contrasteY} shows the counts of synonyms ($n_{syn}$) and antonyms ($n_{ant}$) in explicit DRs of type Contrast and Concession, while Figure \ref{fig:producto-concesion-contrasteX} shows the corresponding counts in implicit Contrast and Concession DRs. In all cases we used the representation $R_\text{all}$.

%%%%%%%%%%%%%%%%%%%%%%%%%%%%%%%%%%%%%%%%%%%%%%%%%%%%%%%%%%%%%%%%%%%%%%%%%%%%%

In order to quantify the differences between the heatmaps, we conducted a non-parametric Mann-Whitney-Wilcoxon test \cite{Neuhauser2011}. Table \ref{tab:significance} summarizes the results.

\begin{table}[H]
    \caption{Statistical significance tests using Mann-Whitney-Wilcoxon to measure differences between the heat maps of synonymy-antonymy correspondences within a discourse relation.}
    \underline{$\hspace{\textwidth}$}
    \renewcommand{\arraystretch}{1.5}
    \begin{tabular}{cll}
    Test & \hspace{5em}Data ($U_1,U_2$) & $p$-value \\ \hline
        1 & Contrast-Explicit, Concession-Explicit & $0.00012$\\ \hline % triangulo
        2 & Contrast-Implicit, Concession-Implicit & $0.35807$ \\ \hline
        3 & Contrast-Explicit, Contrast-Implicit & $0.43556$ \\ \hline
        4 & Concession-Explicit, Concession-Implicit & $4.7934$E-6 \\ \hline % rectángulos
        % Group C & Connectives with less than 14 instances. \\ \hline  % circulo
    \end{tabular}
    \label{tab:significance}
\end{table}

\section{Discussion of the Analysis}    
\label{sec:discusion}

%The present study was motivated by theoretical and methodological questions. 
The first question addressed in this article is a methodological one: How can discourse relations be computationally modeled in order to capture the contribution of the lexical semantics to the discourse relation meaning? We proposed representations of DRs based on POS-bags-of-synonyms/antonyms and were able to computationally capture both, inter-relation semantic patterns (i.e. patterns of synonyms/antonyms found in arguments of the whole set of contrast or concession relations occurrences), and intra-relation patterns (i.e. matches of synonymy/antonymy between $\text{arg}_1$ and $\text{arg}_2$ of a discourse relation).

Although discourse relations have been the subject of growing attention in corpus linguistic studies as well as in NLP and computational linguistics research, the gap between linguistic and computational approaches remains wide. In recent years, we have seen the emergence of ``Transformers'' which are deep neural networks based on self-attentional mechanisms, which have been shown to be able to better deal with long-range correlations in text processing \cite{Vaswani2017}. Prominent state-of-the-art models like ``GPT'' \cite{Radford2019} and ``XLNet'' \cite{yang_xlnet_2019} are pre-trained using autoregressive language models, while ``BERT'' \cite{devlin-etal-2019-bert} uses a denoising approach. The usefulness of these models to discover latent relations between text units or as text analysis tools has been demonstrated in countless contributions for solving complex tasks, such as sentiment analysis \cite{hoang-etal-2019-aspect},  semantic textual matching \cite{Tingyu2021}, and semantic role labelling \cite{larionov-etal-2019-semantic} among many others.

%\cite[e.g. see][]{hoang-etal-2019-aspect,Tingyu2021,larionov-etal-2019-semantic}. 
Still, the question remains as to what exactly these models extract at the linguistic level, or in other words, what these models tell us about the properties of language.
With respect to this question, a recent analysis of BERT \cite{rogers-etal-2020-primer} shows that this model can be very useful for extracting some linguistic knowledge, especially syntactic, and for producing embedded representations with high quality contextualized distributional properties. Despite this, BERT remains vulnerable to variations in context \cite{atwell2021we} or syntactic structure \cite{rogers-etal-2020-primer}. On the other hand, extracting knowledge about the linguistic functions that BERT attention heads manage to classify is an arduous and complex process, and the role of attention remains moot \cite{rogers-etal-2020-primer}. Although we know that these models are capable of establishing long-range correlations in the presence of a very large amount of data, we only have some hints about how the model makes decisions based on how its different processing layers ``pay attention'' to some words or tokens. 

In view of this, we believe that our proposal, although simple in its structure, enables a corpus analysis that can lead to a more detailed, cautious and eventually basic, or elementary, interpretation of how the relations of synonymy and antonymy, which exist in concession and contrast discourse relations, come into play and determine certain properties attributed to this type of linguistic discourse relations. In this sense, our proposed representations of discourse relations, based on POS-bags-of-synonyms/antonyms, allows the computational study of these linguistic forms in a more transparent and linguistically meaningful way. 

%%%%%%%%%%%%%%%%%%%%%%%%%%%%%%%%%%%%%%%%%%%

The quantitative analysis of the graphs resulting from our representations $R_\text{all}$, $R_{\text{all}-\text{aj}}$, $R_{\text{all}-\text{av}}$, $R_{\text{all}-\text{v}}$ and $R_{\text{all}-\text{n}}$ provides insights into the inter-relation patterns of synonymy and antonymy. Firstly, recall that the measure we proposed ($\varphi$, cf. Eq. \ref{eq:phi}) indicates whether there is a node that dominates the network to a large extent, attracting, or connecting with, many more nodes than the rest of the nodes on average. This means that our representations allow the emergence of arguments that are reduced to a single node, which concentrates a significant number of connections to other arguments. This concentrating node synthesizes many others: all the arguments represented in that node are identical in terms of the lexical-semantic content of our sets. In linguistic terms, this indicates that there are lexical-semantic patterns found across contrast and across concession relations in the corpus; in other words, although these representations are mainly built to provide answers to our theoretical questions ---more directly pertaining to intra-relation patterns---, the proposed representations also offer a measure of the topic homogeneity of the corpus and the distribution of the lexical patterns of synonymy and antonymy across discourse relations: the higher the number of arguments concentrating in a single node, the higher the semantic homogeneity of the content included in discourse relation segments in the corpus. In this sense, it is interesting to note that the value of $\varphi$ when all parts of speech are included is slightly higher in concession (solid blue bars in Figures \ref{fig:phi_all_vs_no-aj} to \ref{fig:phi_all_vs_no-n}) than in contrast relations (solid orange bars in Figures \ref{fig:phi_all_vs_no-aj} to \ref{fig:phi_all_vs_no-n}). This indicates that, in both types of discourse relations, there is a significant number of argument representations with central roles, although this phenomenon is more frequent in concession relations. 

Our analysis provides answers to the three theoretical questions posed in the study. We first wondered how much different parts of speech contribute to the representation of contrast and concession discourse relations. Previous studies considering lexical terms expressing opposition either limit themselves to one part of speech \cite{spenader2007antonymy}, do not specify the word class of the lexical elements considered or do not analyze the contribution of the different parts of speech \cite{feltracco2018lexical,marcu2002unsupervised}. Our results show that the inclusion or exclusion of all the parts of speech in the representation plays a relevant role: removing adjectives, adverbs, verbs and nouns results in changes in the graph complexity, revealing that all of them contribute to capturing inter-relation lexical patterns between arguments. Their contribution in contrast and concession relations is, nevertheless, different. 

In contrast relations, removing nouns, verbs, adjectives and, to a lesser degree, adverbs from the representation results in less complex graphs (less nodes with several branches), indicating that the synonymy and antonymy of all parts of speech are productively contributing to grouping together the representation of arguments in contrast relations. Topic homogeneity in contrast relations is best captured with representations that include synonymy and antonymy of  nouns (responsible for topic continuity), verbs and adjectives (coding predicates) and, to a lesser degree, adverbs. On the other hand, in concession relations (group A), we find the opposite pattern:  removing adjectives, adverbs, verbs and, to a lesser degree, nouns, results in more complex graphs. When all parts of speech are included in the representation, there are less discourse segments collapsing in the same representation than when one part of speech is removed. This means that, although our representations capture inter-relation synonymy/antonymy patterns between arguments in concession relations, they also reveal that many of the verbs, adjectives and adverbs included in our sets are in fact discriminating between arguments, in such a way that when they are removed from the representation, the complexity of the graph increases. In turn, when nouns are removed from concession representations, the change is minor, indicating that nouns in the concession representations are not playing such a discriminating role and the topic homogeneity of the concession segments of the corpus is more robustly captured by means of nouns, encoding referents being talked about in the text.

%\jhv{¿Esta discusión es suficiente para responder a la pregunta acerca del rol de las POS en las relaciones discursivas?}

The remaining two questions are answered through the analysis of intra-relation synonymy/antonymy patterns. The first question was whether contrast and concession discourse relations were differentiated using this model of representation and, specifically, whether the presence of antonymy and synonymy patterns between argument 1 and 2 of a given relation differed in contrast and concession DRs, as captured by our representations. First, notice that the presence of synonymy and antonymy between the two arguments in contrast and concession relations is, overall,  high (around 70\%), in contrast with the findings in previous studies using other corpora \cite{feltracco2018lexical,spenader2007antonymy}.

Overall, the presence of pairs of synonyms and antonyms between arguments is almost parallel in the two types of discourse relations. However, from Table \ref{tab:significance}, we see that the difference between synonymy-antonymy relations within Explicit-Contrast and Explicit-Concession is statistically significant with a p-value below 0.1\%: concession relations show significantly more antonymy and synonymy counts than contrast. This is not the case regarding the difference between synonymy-antonymy relations within implicit DRs of type Contrast and Concession. When antonyms and synonyms appear in our data, antonymy is, overall, less frequent than synonymy in both contrast and concession relations. 
The presence of more intra-relation lexical matches in explicit concession relations than in contrast relations is, in principle, unexpected, since contrast discourse relations were expected to be more dependent on the lexical-semantic content of its arguments than concession relations (see Section 2). This result might be due to the PDTB3 encoding procedure, in which “whenever \textsc{Concession} can be taken as holding, it is annotated as such, even if \textsc{Contrast} also holds by definition” \cite[cf.][p. 24]{webber2019penn}. This causes that a discourse relation in which there are at least two differences between $\text{arg}_1$ and $\text{arg}_2$ is not necessarily marked as \textsc{Contrast}; instead, it could be tagged as \textsc{Concession}. On a closer look, it is, however, interesting to notice that antonymy is in fact slightly more frequent in explicit contrast than in explicit concession relations, a tendency in agreement with previous studies \cite{feltracco2018lexical,crible2022syntax}. Regarding synonymy,  we posit that intra-relation synonymy is overall contributing to creating coherence through topic continuity \cite{lei2018linguistic}, a discourse function equally displayed in contrast and concession relations. 

Finally, we wondered whether implicit and explicit discourse relations behave similarly in terms of this representation. The literature on discourse relations modeling seems to operate under the assumption that explicit and implicit discourse relations follow the same linguistic patterns, thus training their models with explicit discourse relations in order to infer implicit ones. This assumption, nevertheless, is at odds with the more straightforward communicative hypothesis that the speaker takes into account the difficulty of inferring the discourse relation in order to decide between the explicit or implicit connective \cite{asr2012implicitness}. This idea has mostly been put forward in order to compare the use of connectives for more basic (additive and positive causal) versus less basic (negative causal) discourse relations \cite{das2019multiple,hoek2015factors}. Under the same logic, if a discourse relation can be easily inferred from the explicit semantic content in the two arguments, implicit connectives would be expected, whereas explicit ones would be more frequent when the conceptual semantics contributes to a lesser extent to establish the discourse relation.

With this hypothesis, the analysis of synonymy and antonymy global match counts between first and second arguments allows us to compare implicit and explicit contrast and concession relations. The results indicate that there is a significant difference between the presence of synonymy-antonymy matches in explicit concession DRs vs. implicit concession DRs: explicit discourse relations show a higher proportion of antonymy-synonymy matches than implicit ones. The difference between explicit and implicit contrast relations is not significant. In our data, therefore, when the lexical conceptual semantics in the arguments contributes with more information (mainly in terms of synonymy) to establishing the contrast or concession discourse relation, it is not more likely for the connective to remain implicit. In fact, the opposite occurs for concession relations. It is possible to think that the concession relations that writers in our corpus decided to leave implicit were easily inferred based on contextual or discourse knowledge that antonymy and synonymy -and, therefore, our representations- are not capturing, whereas explicit connectives frequently co-occur with intra-relations lexical matches, specially of synonyms, narrowing the kind of discourse relation holding between arguments. 

Lastly, the proposed representations are useful to capture differences among discourse markers. Even though a detailed analysis of the behavior of each connective is out of the scope of this article, the graphs analyzing but, while and still (Section 4.2) are evidence that specific connectives show different patterns regarding the lexical semantic matches in intra- and inter-discourse relations. Previous literature has addressed the idea that discourse markers can be organized in terms of their “cue strength” \cite{asr2012measuring}, a probabilistic measure of their ambiguity and monosemy, and suggests that the way different discourse markers interact with signals in their context is related to their strength or weakness \cite{crible2022syntax}. In this line, our proposal opens the possibility to further analyze how different connectives interact with lexical semantic information in the discourse context.

\section{Conclusion}    
\label{sec:conclusion}

Our work addresses important questions in corpus linguistics regarding the role of semantic signals in contrast and concession discourse relations. To achieve this, we propose a computational modeling approach to discourse relations in corpus that is transparent about how certain semantic features are automatically analyzed for the sake of linguistic interpretability of the results. In this sense, our approach allows us to abstract lexical-semantic signals of synonymy and antonymy between the arguments of each discourse relation, thus obtaining information regarding the contribution of synonymy and antonymy in the signaling of discourse relations and showing the differences and similarities between types of relations (contrast vs. concession, and implicit vs. explicit), according to the different word classes (POS). 

%Our work is a contribution in this direction, proposing 

Our results shed light on possible patterns of co-occurrence of semantic cues and connectives as means of signaling discourse relations.  By proposing a quantitative way of assessing these patterns in large amounts of data in linguistic corpora, our method and results contribute to understanding the extent to which these patterns correspond to linguistic-cognitive hypotheses about the use and processing of discourse relations. 

Although discourse relations have been the subject of growing attention in corpus linguistic studies as well as in NLP and computational linguistics research, the gap between linguistic and computational approaches remains wide, and scarce efforts are being made to deepen the dialogue between these disciplines. Bridging this gap, although challenging, is important for interdisciplinary work and offers a promising landscape towards a more complete understanding of discursive linguistic phenomena. We believe our work is a contribution in this direction. 
%Motivated by the idea that the dialogue between computational analysis and corpus and theoretical linguistics can and should be strengthened.
%this article proposed a model to computationally represent discourse relations based on the features of lexical synonymy and antonymy and used this computational analysis to provide answers to linguistic questions through corpus analysis. 
Not the least, we believe our method opens the possibility to extend this kind of analysis to a broader audience, as the methods employed may be automated. Hence, it is possible to extend this kind of analysis to other corpus that may or may not be annotated, contributing thus to deepen research in corpus linguistics.

\section*{Acknowledgements}
Research was partially funded by CONACYT Project A1-S-24213 of Basic Science and CONACYT grants 28268, 29943 and 732458. The authors thank CONACYT for the computer resources provided through the INAOE Supercomputing Laboratory's Deep Learning Platform for Language Technologies.

Conflicts of Interest: The authors declare no conflict of interest. The funders had no role in the design of the study; in the collection, analyzes, or interpretation of data; in the writing of the manuscript; or in the decision to publish the results.

\bibliographystyle{unsrt} 
\bibliography{references}

\end{document}